\pdfoutput=1

\documentclass[11pt, dvipsnames]{article}

\usepackage{authblk}
\usepackage[]{acl}

\usepackage{times}
\usepackage{latexsym}

\usepackage[T1]{fontenc}

\usepackage[utf8]{inputenc}

\usepackage{microtype}

\usepackage{inconsolata}

\usepackage{amsmath}
\usepackage{graphicx}
\usepackage{float}
\usepackage{cleveref}
\usepackage{setspace}
\usepackage{booktabs}
\usepackage{siunitx}
\usepackage[position=top]{subfig}
\usepackage{chngcntr}
\usepackage{multirow}
\usepackage{xspace}
\usepackage{colortbl}
\usepackage{booktabs}
\usepackage{wrapfig}
\usepackage{stfloats}

\newcommand{\FINAL}[1]{}

\newcommand{\datasetAcronym}{HQP}
\newcommand{\datasetTextOnly}{\textsf{\datasetAcronym}\xspace}
\newcommand{\dataset}{\textsf{\datasetAcronym}\xspace}

\newcommand{\datasetplusAcronym}{HQP+}
\newcommand{\datasetplusTextOnly}{\textsf{\datasetplusAcronym}\xspace}
\newcommand{\datasetplus}{\textsf{\datasetplusAcronym}\xspace}

\newcommand{\binarylabel}{binary label\xspace}
\newcommand{\binarylabelshort}{\emph{BL}\xspace}

\newcommand{\strategylabel}{propaganda-stance label\xspace}
\newcommand{\strategylabelshort}{\emph{PSL}\xspace}

\usepackage{tikz}
\definecolor{powerpointgreen}{RGB}{197, 224, 180}
\definecolor{powerpointblue}{RGB}{132,151,176}
\definecolor{powerpointorange}{RGB}{244,177,131}
\newcommand*\circledblue[1]{\tikz[baseline=(char.base)]{
            \node[shape=circle,draw=powerpointblue!150,fill=powerpointblue,thick,inner sep=1pt] (char) {\small#1};}}
\newcommand*\circledorange[1]{\tikz[baseline=(char.base)]{
            \node[shape=circle,draw=powerpointorange!150,fill=powerpointorange,thick,inner sep=1pt] (char) {\small#1};}}
\newcommand*\circledgreen[1]{\tikz[baseline=(char.base)]{
            \node[shape=circle,draw=powerpointgreen!150,fill=powerpointgreen,thick,inner sep=1pt] (char) {\small#1};}}

\usepackage{array}
\newcolumntype{P}[1]{>{\centering\arraybackslash}p{#1}}

\usepackage{pifont}

\newcommand{\cmark}{\textcolor{ForestGreen}{\ding{51}}}%
\newcommand{\xmark}{\textcolor{BrickRed}{\ding{55}}}%

\newcolumntype{R}[1]{>{\raggedleft\let\newline\\\arraybackslash\hspace{0pt}}m{#1}}

\title{\datasetTextOnly: A Human-Annotated Dataset for Detecting Online Propaganda}

\makeatletter
\newcommand\email[2][]%
   {\newaffiltrue\let\AB@blk@and\AB@pand
      \if\relax#1\relax\def\AB@note{\AB@thenote}\else\def\AB@note{\relax}%
        \setcounter{Maxaffil}{0}\fi
      \begingroup
        \let\protect\@unexpandable@protect
        \def\thanks{\protect\thanks}\def\footnote{\protect\footnote}%
        \@temptokena=\expandafter{\AB@authors}%
        {\def\\{\protect\\\protect\Affilfont}\xdef\AB@temp{#2}}%
         \xdef\AB@authors{\the\@temptokena\AB@las\AB@au@str
         \protect\\[\affilsep]\protect\Affilfont\AB@temp}%
         \gdef\AB@las{}\gdef\AB@au@str{}%
        {\def\\{, \ignorespaces}\xdef\AB@temp{#2}}%
        \@temptokena=\expandafter{\AB@affillist}%
        \xdef\AB@affillist{\the\@temptokena \AB@affilsep
          \AB@affilnote{}\protect\Affilfont\AB@temp}%
      \endgroup
       \let\AB@affilsep\AB@affilsepx
}
\makeatother

\author[*]{\textbf{Abdurahman Maarouf}}
\author[*]{\textbf{Dominik Bär}}
\author[*]{\textbf{Dominique Geissler}}
\author[*]{\textbf{Stefan Feuerriegel}}
\affil[*]{Munich Center for Machine Learning (MCML) 
  \& LMU Munich}
\email{\{a.maarouf, baer, d.geissler, feuerriegel\}@lmu.de}

\begin{document}
\maketitle
\begin{abstract}
Online propaganda poses a severe threat to the integrity of societies. However, existing datasets for detecting online propaganda have a key limitation: they were annotated using \emph{weak} labels that can be noisy and even incorrect. To address this limitation, our work makes the following contributions: (1)~We present \dataset: a novel dataset ($N=\num{30000}$) for detecting online propaganda with \emph{high-quality} labels. To the best of our knowledge, \dataset is the first large-scale dataset for detecting online propaganda that was created through human annotation. (2)~We show empirically that state-of-the-art language models fail in detecting online propaganda when trained with weak labels (AUC: 64.03). In contrast, state-of-the-art language models can accurately detect online propaganda when trained with our high-quality labels (AUC: 92.25), which is an improvement of $\sim$44\%. (3)~We show that prompt-based learning using a small sample of high-quality labels can still achieve a reasonable performance (AUC: 80.27) while significantly reducing the cost of labeling. (4)~We extend \dataset to \datasetplus to test how well propaganda across different contexts can be detected. Crucially, our work highlights the importance of high-quality labels for sensitive NLP tasks such as propaganda detection.
\end{abstract}

\noindent {\footnotesize\textbf{Disclaimer:} Our work contains potentially offensive language and manipulative content. Reader's discretion is advised.}

\section{Introduction}


Propaganda is used to influence, persuade, or manipulate public opinions \cite{Smith.2022}. Nowadays, propaganda is widely shared on social media as a practice of modern warfare (e.g., in the ongoing Russo-Ukrainian war) and thus poses a significant threat to the integrity of societies \cite{Kowalski.2022}. 


\begin{table}[t!]
\centering
\tiny
\begin{tabular}{p{0.05cm}p{0.02cm}p{2.2cm}R{1.0cm}R{1.3cm}R{0.65cm}}
\toprule
\textbf{\datasetTextOnly} &                   &           & Propaganda & No propaganda & Overall  \\
\midrule
\multirow{3}{*}{\protect\circledblue{R}} & \multirow{3}{*}{\huge\{}   & Num. of posts ($N$)                     & \num{4610}               & \num{25390}    & \num{30000}              \\
&& Avg. post length (in chars) & 238.71             & 216.90    &   220.25           \\
&& Num. of unique authors                     & \num{3910}               & \num{20140}    & \num{23317}       \\    
\midrule
\textbf{\datasetplusTextOnly}    &                &           & Propaganda & No propaganda & Overall  \\
\midrule
\multirow{3}{*}{\protect\circledblue{R}} & \multirow{3}{*}{\huge\{} & Num. of posts ($N$)                     & \num{4610}               & \num{25390}    & \num{30000}              \\
&& Avg. post length (in chars) & 238.71             & 216.90    &   220.25           \\
&& Num. of unique authors                     & \num{3910}               & \num{20140}    & \num{23317}       \\    
\addlinespace
\multirow{3}{*}{\protect\circledorange{M}} & \multirow{3}{*}{\huge\{} & Num. of posts ($N$)                     & \num{337}               & \num{663}    & \num{1000}              \\
&& Avg. post length (in chars) & 231.63             & 225.23    &   227.38           \\
&& Num. of unique authors                     & \num{290}               & \num{559}    & \num{789}       \\    
\addlinespace
\multirow{3}{*}{\protect\circledgreen{U}} & \multirow{3}{*}{\huge\{} & Num. of posts ($N$)                     & \num{256}               & \num{744}    & \num{1000}              \\
&& Avg. post length (in chars) & 231.65             & 229.35    &   229.94           \\
&& Num. of unique authors                     & \num{198}               & \num{609}    & \num{769}       \\    
\bottomrule
\end{tabular}
\caption{Summary statistics. \dataset contains \protect\circledblue{R} Russian propaganda. \datasetplus extends it with \protect\circledorange{M} anti-Muslim propaganda in India, and \protect\circledgreen{U} anti-Uyghur propaganda.}
\label{tab:summary_stats}
\vspace{-0.5cm}
\end{table}


Existing NLP works for propaganda detection \citep[e.g.,][]{Rashkin.2017, BarronCedeno.2019, DaSanMartino.2019, Wang.2020, Vijayaraghavan.2022} rely upon datasets that were exclusively annotated using \emph{weak} labels and were, therefore, \underline{not} validated by humans. Because of this, labels can be noisy and even incorrect. We later provide empirical support for this claim and show that the overlap between weak labels and human annotations is only $\sim$41\%. 




To fill this gap, we develop \dataset: a novel dataset with \textbf{h}igh-\textbf{q}uality labels for detecting online \textbf{p}ropaganda. \dataset consists of $N=\num{30000}$ posts in English from the Russo-Ukrainian war (\Cref{tab:summary_stats}). We use human annotation and validation to generate \emph{high-quality} labels. We then leverage state-of-the-art, pre-trained language models (PLMs), i.e., BERT, RoBERTa, and BERTweet, to benchmark the performance in detecting online propaganda using weak labels vs. our high-quality labels. We find that high-quality labels are crucial for detecting online propaganda. We further acknowledge that human annotation also incurs labeling costs, and, to address this, we extend our work to few-shot learning (i.e., prompt-based learning). We further introduce \datasetplus ($N_+=\num{32000}$), an extended version of \dataset that adds two additional contexts: anti-Muslim ($\num{1000}$ posts) and anti-Uyghur propaganda ($\num{1000}$ posts). 



Our main \textbf{contributions} are as follows:\footnote{Code and data are available via \url{https://github.com/abdumaa/HiQualProp/}.}
\begin{enumerate}
\item We construct \dataset, a novel dataset with high-quality labels for online propaganda detection using human-annotated labels. 
\item We show that PLMs for detecting online propaganda using high-quality labels outperform PLMs using weak labels by a large margin.
\item We adapt few-shot learning for online propaganda detection by prompting PLMs.
\item We extend \dataset to \datasetplus to test the ability of cross-context propaganda detection.
\end{enumerate}

\section{Related Work} \label{sec:rw}


\textbf{Detecting harmful content:} Prior literature in NLP has aimed to detect a broad spectrum of harmful content such as hate speech \citep[e.g.,][]{Badjatiya.2017, Mathew.2021, Pavlopoulos.2022}, rumors \citep[e.g.,][]{Zhou.2019, Bian.2020, Xia.2020, Wei.2021}, and fake news \citep[e.g.,][]{Zellers.2019, Liu.2020, Lu.2020, Jin.2022}. Further, claim detection has been studied, for example, in the context of the Russo-Ukrainian war \cite{LaGatta.2023}. Overall, literature for detecting harmful content makes widespread use of datasets that were created through human annotations \citep[e.g.,][]{Founta.2018, Thorne.2018, Mathew.2021}. 


\textbf{Detecting propaganda content:} Previous works for propaganda detection can be loosely grouped by the underlying content, namely (1)~official news and (2) social media. 


(1)~\emph{News}. To detect propaganda in official news, existing works leverage datasets that originate from propagandistic and non-propagandistic news outlets \citep{Rashkin.2017, BarronCedeno.2019, DaSanMartino.2019, DaSanMartino.2020, Solopova.2023}, yet these datasets are not tailored to online content from social media. As a case in point, \citet{Wang.2020} found challenges in the capability of machine learning to transfer propaganda detection between news and online content. 

(2)~\emph{Social media.} To detect propaganda in social media, existing works create datasets from online platforms such as Twitter/X. For example, source-based datasets \citep[e.g.,][]{Wang.2020, Guo.2022} combine a random sample of posts (for the non-propagandistic class) with a sample of posts from propagandistic sources (for the propagandistic class). However, source-based datasets rely on the source and not on the content for annotation. TWEETSPIN \cite{Vijayaraghavan.2022} collects posts that are annotated with weak labels along different types of propaganda techniques by mining accusations in the replies to each post. While human-annotated datasets for detecting propaganda on social media exist \citep{Dimitrov.2021, Moral.2023}, they are not large-scale. Notably, all existing large-scale datasets for detecting online propaganda were created through weak annotation (see Table~\ref{tab:existing-datasets}). To this end, labels can oftentimes be noisy or even incorrect. We develop a new, large-scale human-annotated dataset with \emph{high-quality} labels and provide a rigorous comparison of weak annotation vs. human annotation for propaganda detection on social media. 


\textbf{Few-shot learning in NLP:} Generally, constructing large-scale datasets with high-quality labels in NLP is costly. Hence, there is a growing interest in few-shot learning. Common methods typically leverage prompting, where the downstream task is reformulated to resemble the masked language modeling task the PLM was trained on \citep[e.g.,][]{Radford.2019, Brown.2020, Gao.2021, Schick.2021, Liu.2023}. Prompting has been highly successful in few-shot learning, e.g., for rumor detection \cite{Lin.2023} or humor detection \cite{Li.2023}. However, to the best of our knowledge, no work has so far adapted few-shot learning to detect propaganda.

\begin{table*}[h!]
    \begin{center}
        \tiny
                \begin{tabular}{l llclc}
                    \toprule
                    Dataset & Domain &  Level & Human ann. & Model & Few-shot \\
                    \midrule
                    \citet{Rashkin.2017} & News & Document & \xmark & LSTM & \xmark \\
                    \citet{BarronCedeno.2019} & News & Document & \xmark & Maximum entropy classifier & \xmark \\
                    \citet{DaSanMartino.2019} & News & Fragment & \cmark & Multi-granularity network & \xmark \\
                    \citet{Solopova.2023} & News & Document & \xmark & BERT & \xmark \\
                    \midrule
                    \citet{Wang.2020} (``TWE'') & Social media & Short-text & \xmark & LSTM & \xmark \\
                    \citet{Vijayaraghavan.2022} (``TWEETSPIN'') & Social media & Short-text & \xmark & Multi-view transformer & \xmark \\
                    \midrule
                    \textbf{\datasetTextOnly (ours)} & Social media & Short-text & \cmark & BERT, RoBERTa, BERTweet & \cmark \\
                    \midrule
                \end{tabular}
    \end{center}
    \vspace{-0.3cm}
    \caption{Overview of existing large-scale datasets for propaganda detection aimed at (i)~news and (ii)~online content.}
    \label{tab:existing-datasets}
    \vspace{-0.5cm}
\end{table*}

\section{Dataset Construction}\label{sec:data_construction}


We construct a human-annotated dataset of English social media content with propaganda (\dataset). For this, we construct a corpus of posts with Russian propaganda from the 2022 invasion of Ukraine. We collect posts from February 2021 until October 2022, i.e., our timeframe starts one year before the invasion due to the widespread opinion that the invasion was planned far in advance.\footnote{\url{https://www.nytimes.com/2021/04/09/world/europe/russia-ukraine-war-troops-intervention.html}} We intentionally choose the Russo-Ukrainian war due to its significance for global politics \cite{Kowalski.2022} and the size of the propaganda campaign \cite{Geissler.2022}. 


Our methodology for constructing \dataset (and \datasetplus) follows best practices for human annotation \cite{Song.2020}. Specifically, we follow a three-step process: (1)~data collection, (2)~sampling, and (3)~human annotation.


\subsection{Data Collection}
\label{sec:data_collection}

\begin{figure}[t]
    \captionsetup{position=top}
    \centering
    \includegraphics[width=0.9\linewidth]{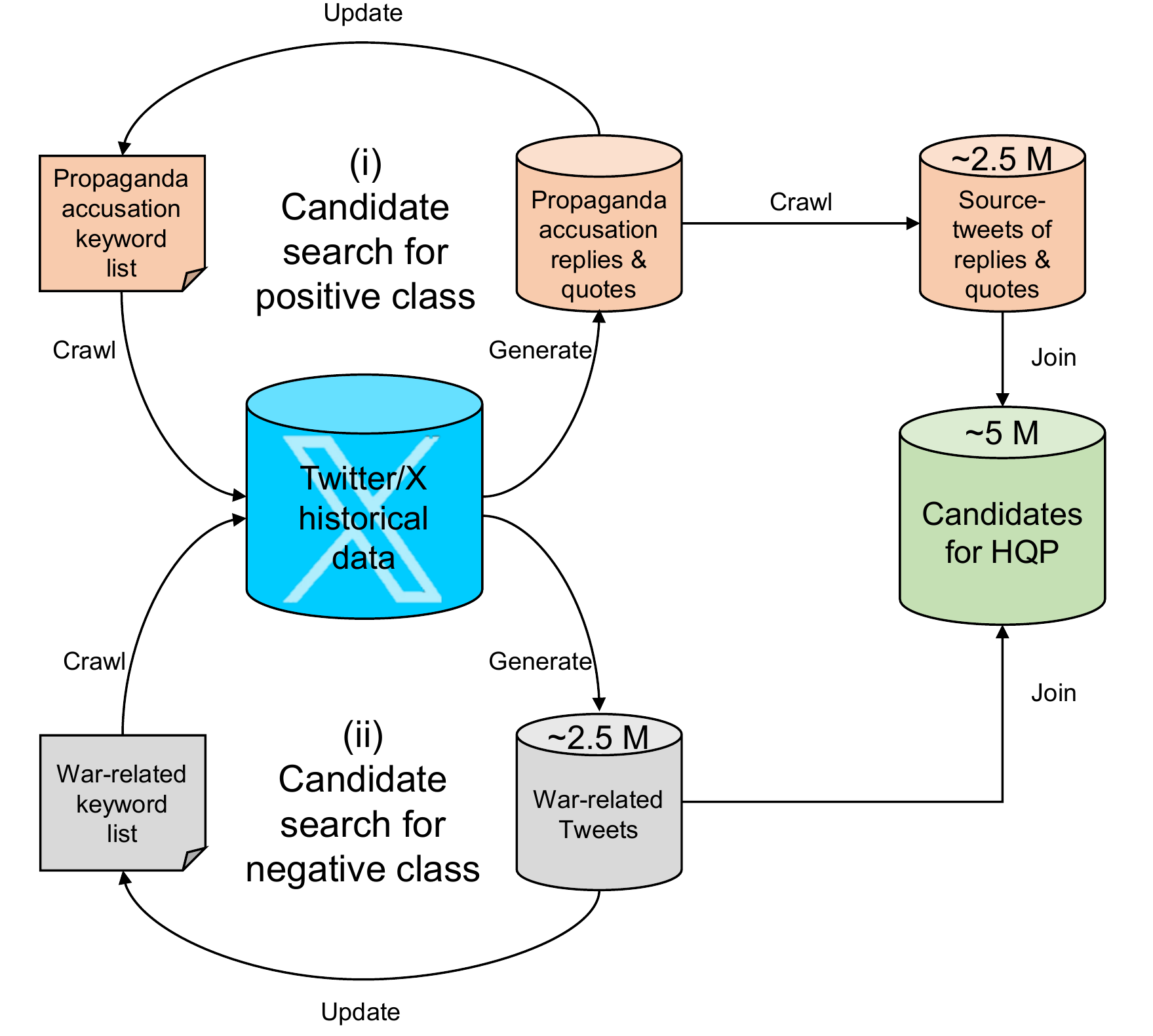}
    \caption{Data collection of candidate posts for \dataset.}
\label{fig:datageneration}
\vspace{-.7cm}
\end{figure}

 
Social media content with propaganda is rare in comparison to non-propaganda (i.e., well below 0.1\%). Therefore, simply collecting a random subset of posts will contain only very few samples from the positive class (i.e., propaganda). Instead, we follow the methodology in \citet{Founta.2018} and perform a stratified search. Thereby, we separately generate candidates for the positive class ($D_{+}$) and for the negative class ($D_{-}$) as shown in \Cref{fig:datageneration}. In $D_{-}$ we collect context-related samples that discuss topics related to the Russo-Ukrainian war but are likely \emph{not} propaganda. Thereby, we create a challenging setting in which we can evaluate how accurately propaganda and non-propaganda can be discriminated.

\textbf{(i) Candidate search for positive class ($D_+$):} Analogous to \citet{Vijayaraghavan.2022}, we expect that propaganda on Twitter/X is often called out in replies or quotes (e.g., some users debunk propaganda). We thus access the Twitter Historical API and perform a keyword-based search. Specifically, we crawl replies and quotes that contain phrases (keywords) that may accuse the original post of propaganda, such as (``\texttt{russian}'' $\land$ ``\texttt{propaganda}'') or (``\texttt{war}'' $\land$ ``\texttt{propaganda}''). The full list is in \Cref{tab:accusation_phrases_R} in the supplements. 

We create the list of search keywords through an iterative process: (1)~In each iteration, the current list of keywords is used to filter for English-language replies and quotes from Twitter/X. (2)~We then manually scan the most frequent words (including bi- and tri-grams) for phrases that can potentially qualify as propaganda accusations. (3)~We add these to our list of keywords. We repeat the process for three iterations and use the final list of search keywords to retrieve our set of replies and quotes. Afterward, we crawl the corresponding source posts, which resulted in $\sim$2.5 million candidates for $D_+$.

\textbf{(ii) Candidate search for negative class ($D_{-}$):} To collect candidates for the negative class, we crawl a random sample of 2.5 million posts that discuss the Russo-Ukrainian war but that have not necessarily been identified as propaganda through users. For this, we use a similar iterative procedure to generate a keyword list as for the positive class. However, we now perform a keyword search only for source posts (but not for replies or quotes). Example keywords are (``\texttt{russia}'' $\land$ ``\texttt{war}'') and (``\texttt{ukraine}'' $\land$ ``\texttt{war}''). The complete list of search keywords is in Table~\ref{tab:warrelated_phrases_R} in the supplements.

\textbf{Postprocessing:} We postprocess the candidates for both the positive class ($D_+$) and the negative class ($D_{-}$). Specifically, we filter out duplicates, non-English posts, and very short posts (i.e., fewer than 5 words). The resulting union of both postprocessed candidate sets contains  $|D_+| + |D_-|\approx$ 3.2 million samples. 

Unlike \citet{Vijayaraghavan.2022}, we do not perform weak labeling by simply assigning a label to a post depending on whether it is in $D_+$ or $D_-$, respectively. Instead, we generate high-quality labels through human annotation. This is motivated by our observation that many samples in $D_+$ cover the Russo-Ukrainian war but do not qualify as propaganda. Hence, weak labeling would lead to many false positives.

\subsection{Boosted Sampling}


We collect $N=\num{30000}$ posts from the postprocessed union of $D_+$ and $D_-$ for human annotation. We adopt boosted sampling \cite{Founta.2018} as we observe that the majority of samples from the previous step cover ``normal'' content and do not necessarily qualify as propaganda. We want a sufficient proportion of positive class labels in our dataset, since, otherwise, the dataset will not be useful for the research community. To address the class imbalance, we use weighted sampling. For this, we generate weights $w_i$ for each post as the inverse term frequency of potential propaganda phrases (e.g., ``\texttt{nobody talks about}''), that is, 
\begin{align}\label{eq:signal-weighted-sampling}
    w_i = & \, \frac{n_i}{\sum_{j=1}^{M} n_j} ,
\end{align}
where $n_i$ is the number of occurrences of propaganda phrases and $M = |D_+| + |D_-|$. We use the list of 189 potential propaganda phrases from \citet{Vijayaraghavan.2022}. As a result, our boosted sampling approach will increase the likelihood that actual propaganda content (true positives) is later annotated (rather than false positives).

\subsection{Human Annotation}\label{sec:data_annotation}


To annotate our data, we recruit human workers from Prolific (\url{https://www.prolific.co/}). Workers are pre-selected according to strict criteria: residency in the UK/US, English as a first language; enrollment in an undergraduate, graduate, or doctoral degree; a minimum approval rate of 95\%; and at least 500 completed submissions on Prolific.

We follow prior research on propaganda detection \citep[e.g.,][]{BarronCedeno.2019, DaSanMartino.2019, DaMartino.2020, Salman.2023} and use the following definition of propaganda from \citet{Miller.1939} for annotation: ``Propaganda is expression of opinion or action by individuals or groups deliberately designed to influence opinions or actions of other individuals or groups with reference to predetermined ends.'' To further guide workers during the annotation, we provide them with detailed, context-specific guidelines (see \Cref{appendix:annotation_instructions}). The guidelines list relevant news coverage and literature specific to the context. Moreover, we provide context-specific subgroups of content and whether they should be regarded as propagandistic or not. For example, misinformation that favors the Russian government or pro-Russian slurs/slogans in the context of the Russo-Ukrainian war should be regarded as propagandistic.

Given the complexity and subjective nature of propaganda, we put great emphasis on providing meaningful and precise instructions for annotation (see \Cref{appendix:annotation_instructions}). Therefore, the final instructions are a result of several iterations of improvement, each followed by an internal discussion and analysis. In each iteration, we had three workers from Prolific annotate a random sample of 100 posts using the current version of the instructions. We analyzed the resulting annotations and focused on the samples with disagreement. We then aimed to address the corresponding issues in the next update of our instructions. In each iteration, we calculated the inter-annotator agreement using Krippendorff's alpha \cite{Krippendorff.2016} and stopped iterating after surpassing an agreement of 0.8 for the first time. As a result, we updated the instructions three times.


Workers are asked to annotate two labels for each post. The first is a \binarylabel (\binarylabelshort) to classify propagandistic vs. non-propagandistic content. The second is a \strategylabel (\strategylabelshort) aimed at capturing the context-related stance behind propagandistic posts. Therefore, if a post is annotated as propagandistic, the worker is asked to select one of four context-related propaganda stances that are represented in this post (thus giving \strategylabelshort). The four propaganda stances were carefully chosen after manually studying a sample of \num{2000} posts and discussing different options with an expert team of propaganda researchers. Specifically, for \strategylabelshort, workers have to decide whether the propagandistic post has a stance (1)~against the main opposition (e.g., against Ukraine), (2)~pro own stance (e.g., pro Russian government), (3)~against other oppositions (e.g., against Western countries), or (4)~other. Hence, \emph{only} propagandistic posts (\binarylabelshort $=1$) receive one of the four stances. In \Cref{tab:exampleposts_r}, we list five example posts and their corresponding labels, i.e., \binarylabelshort and \strategylabelshort.


Our annotation follows a batch procedure according to best practices \cite{Song.2020}, i.e., a pool of workers annotates a subset of the data to avoid fatigue. We thus split the dataset ($N=\num{30,000}$) into 300 batches with 100 posts each. Each batch is annotated by two workers. Beforehand, we manually annotate 10 posts of each batch with respect to \binarylabelshort to measure the quality of the annotations:\footnote{The internally annotated tweets only serve as quality checks and do not determine the final labels.} If (a)~a worker incorrectly labeled more than 20\% of the internally annotated posts or (b)~the inter-annotator agreement between both workers has a Cohen's kappa \cite{Cohen.1960} $\leq 0.4$, we discard the annotation and repeat the annotation for the batch. Overall, we had to discard and redo 7.5\% of the batch annotations. 

When annotators disagreed on \binarylabelshort for individual posts, we resolve the conflicts as follows: the \binarylabelshort is then re-annotated by randomly assigning it to one of the top 25 annotators. If there is disagreement on the \strategylabelshort after resolving the disagreement on the \binarylabelshort, the final \strategylabelshort is decided by the author team. The latter was the case for only 2.6\% of the posts. Overall, the agreement with the internally annotated posts amounts to $91.92$\% and the inter-annotator agreement (Cohen's kappa \cite{Cohen.1960}) is $0.71$ for \binarylabelshort and $0.62$ for \strategylabelshort. Altogether, this corroborates the reliability of our multi-annotator, multi-batch annotation procedure. \Cref{tab:exampleposts_r} lists five example tweets and their corresponding labels, i.e., \binarylabelshort and \strategylabelshort.

\begin{table*}[h!]
\centering
\scriptsize
\renewcommand{\arraystretch}{1.3}
\begin{tabular}{p{12.9cm}P{0.5cm}p{1.42cm}}
\toprule
Posts                                                                                                                                                                                                                                                                                             & \binarylabelshort & \strategylabelshort                        \\
\midrule
``\texttt{STOP RUSSIAN AGGRESSION AGAINST \#UKRAINE . @USER CLOSE THE SKY OVER UKRAINE ! EXCLUDE RUSSIA FROM THE @USER SECURITY COUNCIL ! \#StopPutin \#StopRussia HTTPURL}''                                                                                                                                  & False  & ---          \\
``\texttt{The Textile Worker microdistrict in Donetsk came under fire ! The Ukraine nazis dealt another blow to the residential quarter At least four civilians were killed on the spot . \#UkraineRussiaWar \#UkraineNazis \#ZelenskyWarCriminal @USER @USER HTTPURL}''                                       & True  & Against main opposition            \\
``\texttt{The denazification of Ukraine continues . In Kherson , employees of the Russian Guard detained two accomplices of the Nazis . During the operation , the National Guard officers detained several leaders of neo-Nazi formations and accomplices of the SBU . HTTPURL HTTPURL}''                     & True  & Pro own stance     \\
``\texttt{Western `leaders' continue with their irrational drive toward WWIII . NATO is a criminal enterprise , an instrument of white power  threat to global humanity . Join anti-NATO protests around the world . HTTPURL}''                                                                                & True  & Against other oppositions  \\
``\texttt{Chinese and Indian citizens must leave Ukraine because Ukraine is run by the Nazi / Zionist fascists since the coup d'etat of 2014 .}'' & True  & Other  \\
\bottomrule
\end{tabular}
\caption{Example posts in our \dataset dataset. \binarylabelshort is a \binarylabel for whether a post is propaganda or not. \strategylabelshort is the \strategylabel.}
\label{tab:exampleposts_r}
\vspace{-0.5cm}
\end{table*}

\subsection{Data Enrichment}\label{enrichment}

\textbf{Propaganda techniques: }We also provide weak labels for the 18 propaganda techniques defined by \citet{DaSanMartino.2019}. For weak labeling, we use the approach in \cite{Vijayaraghavan.2022} and assign weak labels according to their mapping of propaganda phrases to propaganda techniques. The distribution of propaganda techniques is shown in \Cref{sec:proptechniques}.

\noindent \textbf{Linguistic dimensions: }We further provide the linguistic dimensions (e.g., negative and positive emotions) of each post using the LIWC2015 dictionary \cite{Pennebaker.2015}. We refer to \Cref{appendix:lingdim} for an analysis of the linguistic dimensions.

\noindent \textbf{Author meta information: }We further crawled additional meta information about authors (e.g., number of followers); see Appendix~\ref{sec:exp-mf}. We report summary statistics for meta information about authors in \Cref{tab:sum-stats-af} and \Cref{tab:sum-stats-pt}.

\subsection{\datasetplus}\label{hqpplus}

We also provide an extended version called \datasetplus ($N_+=\num{32000}$) where we include broad coverage of different propaganda contexts. Thereby, we provide an additional dataset to detect context-independent patterns of propaganda. We additionally include (1)~anti-Muslim propaganda in India (\num{1000} posts) and (2)~anti-Uyghur propaganda (\num{1000} posts). Our choice is informed by prior social media research \cite{Oxford.2023}. Both contexts have diverse origins and are considered salient propaganda contexts in current media studies \cite{Oxford.2023}. 

Our annotation follows the same procedure as for \dataset: based on boosted sampling and human annotation (i.e., multi-annotator and multi-batch approach) we generate high-quality labels. As the topics differ, we use different keywords to obtain candidate posts for each class of the two new events. We again refer to \Cref{sec:keywords} for the full list of keywords. \Cref{sec:exampleposts} lists exemplary posts for the two new contexts in \datasetplus.

\section{Methods}\label{sec:methods}


In our experiments, we follow state-of-the-art methods from the literature (see Sec.~\ref{sec:rw}) to ensure the comparability of our results. To this end, we use a binary classification task (propaganda $=1$, otherwise $=0$). Results are reported as the average performance over five separate runs. In each run, we divide the dataset into train (70\%), val (10\%), and test (20\%) using a stratified shuffle split.

\subsection{Fine-Tuning PLMs} \label{sec:finetuning}

\noindent\textbf{PLMs:}
We use the following PLMs for our experiments: \textbf{BERT-large} \cite{Devlin.2018}, \textbf{RoBERTa-large} \cite{Liu.2019}, and \textbf{BERTweet-large} \cite{Nguyen.2020}. The latter uses the pretraining procedure from RoBERTa but is tailored to English Twitter posts to better handle social media content. We report implementation details in \Cref{sec:details-finetuning}.\footnote{We also evaluated the performance of fine-tuning PLMs when incorporating author meta information. Implementation details and results are reported in \Cref{sec:exp-mf}.}

\noindent\textbf{Baselines: }
We compare the fine-tuning procedure on our high-quality labels vs. baselines with weak labels. All evaluations are based on separate test splits of \dataset with human verification.

\noindent$\bullet$\,\textbf{TWE:} We fine-tune on weak labels of the public TWE dataset \cite{Wang.2020} as a baseline for source-based datasets.

\noindent$\bullet$\,\textbf{TWEETSPIN:} The TWEETSPIN dataset with weak labels \cite{Vijayaraghavan.2022} is not public, and we thus replicate the data collection procedure ($N=\num{3223867}$).


\noindent$\bullet$\,\textbf{\dataset-weak:} We construct \dataset-weak using the same posts as in \dataset, but with a weak label instead of our high-quality label. Specifically, we map from our classification into $D_+$ and $D_-$ to generate the weak label. Hence, both \dataset and \dataset-weak include the same posts but only differ in their labeling, ensuring a fair comparison. This allows us to isolate the role of weak vs. high-quality labels.\footnote{We also experimented with a variant of the dataset where we used weak labeling for all $\sim$3.2 million samples but found comparable results.}

\subsection{Prompt-Based Learning}\label{sec:promptlearning}


For few-shot learning, we leverage state-of-the-art prompt-based learning \cite{Liu.2023, Gao.2021}, which requires only a small set of labeled samples and thus reduces annotation costs. Prompt-based learning reformulates the downstream classification task to look more like the masked-language-model task the PLM was trained on. For example, for our task, each input sequence could be appended with a textual prompt, e.g., the propagandistic sequence ``\texttt{Ukraine is full of nazis.}'' is continued with the prompt ``\texttt{This is} $\mathrm{[MASK]}$'' (which gives the so-called template). Given a mapping of predefined label words to each class (via the so-called verbalizer), the masked language model predicts the probabilities of each label word to fill the $\mathrm{[MASK]}$ token and thereby the probabilities of each class. Examples of label words could be ``\texttt{propaganda}'' for the class of propaganda and ``\texttt{true}'' for the class of no propaganda. 

Prompt-based learning introduces the task of prompt engineering, i.e., finding the most suitable template and verbalizer to solve the downstream task. In general, manual prompt engineering can be challenging, especially because the performance in the downstream task depends highly on the prompt \cite{Gao.2021}. Therefore, we rely on automatic template generation and automatic verbalizer generation before performing prompt-based fine-tuning. Specifically, we use a three-step procedure: (i)~finding the best template, (ii)~finding the best verbalizer, and (iii)~prompt-based fine-tuning. Implementation details are provided in \Cref{sec:pbl_appendix}.

\textbf{(i)~Automatic template generation:} Here, we use the LM-BFF procedure from \citet{Gao.2021}. We randomly sample $k'$ positive and $k'$ negative examples for training and validation, which thus requires $k = 4 \times k'$ samples overall. We use the seq2seq PLM T5 \cite{Raffel.2020} to generate template candidates. Given the training example and an initial verbalizer\footnote{This initial verbalizer is only used to generate template candidates. We discard the initial verbalizer for the automatic verbalizer generation in step (ii).}, T5 then generates a candidate template by filling in the missing spans. We use beam search to generate a set of 100 candidate templates. Afterward, we fine-tune each template using the training examples and the downstream PLM. Finally, the best-performing template is chosen based on the performance on the val set.

\textbf{(ii)~Automatic verbalizer generation:} We use the method from \citet{Gao.2021} to generate the verbalizer (i.e., to map predictions to our label classes). For each class, we construct a set of 100 candidate tokens based on the conditional likelihood of the downstream PLM to fill the $\mathrm{[MASK]}$ token using the best-performing template from step (i). These candidates are fine-tuned and re-ranked to find the best candidate for each class with regard to the performance on the val set.

\textbf{(iii)~Prompt-based fine-tuning:} We use the best template from step (i) and the best verbalizer from step (ii) to form our prompt. We fine-tune the downstream PLM with this prompt to create the final model for propaganda detection. We refer to the above model as \textbf{LM-BFF}.\footnote{We also evaluate a set of baselines for our prompt-based learning in \Cref{sec:exp-fs-baselines}. However, the results were not better than those reported in \Cref{sec:exp-fs}.}

\subsection{Extension of LM-BFF to Auxiliary-Task Prompting}

We extend the above LM-BFF procedure for inductive learning and use both the \binarylabelshort and the \strategylabelshort labels during prompting. The rationale behind this is three-fold: (1)~We use information about the propaganda stance and thus richer labels, which may improve performance. (2)~Propaganda can be highly diverse, and, through the use of more granular labels, we can better capture heterogeneity. (3)~The overall sample size remains low with only a minor increase in labeling costs. This is beneficial in practice when newly emerging propaganda narratives must be detected and there are thus only a few available samples. 

To leverage both \binarylabelshort and \strategylabelshort labels, we develop a custom architecture for auxiliary-task prompting, which we refer to as \textbf{LM-BFF-AT}. Specifically, we apply steps (i) to (iii) for our two labels \binarylabelshort and \strategylabelshort, separately. This results in two different fine-tuned versions of the downstream PLM with different templates and verbalizers. To classify a given input text, we fuse verbalizer probabilities for each label into a classification head, which computes the final prediction. For the classification heads, we train an elastic net and a feed-forward neural network with one hidden layer on top of the verbalizer probabilities. The val set is used for hyper-parameter tuning. Hyper-parameter grids for the classification heads are reported in \Cref{sec:hplmbffat}. Note that our LM-BFF-AT approach uses two labels but can easily be generalized to $n$ labels.\footnote{We also evaluated the performance of LM-BFF-AT when additionally incorporating author and pinned-post features. Implementation details and results are in \Cref{sec:exp-mf}.}

\section{Experiments}\label{sec:experiments}

\subsection{Weak Labels $\mathbf{\neq}$ High-Quality Labels}

\textbf{RQ1:} \emph{What is the discrepancy between weak labeling vs. human annotation?}

\begin{wrapfigure}{r}{0.25\textwidth}
    \centering
    \includegraphics[width=1\linewidth]{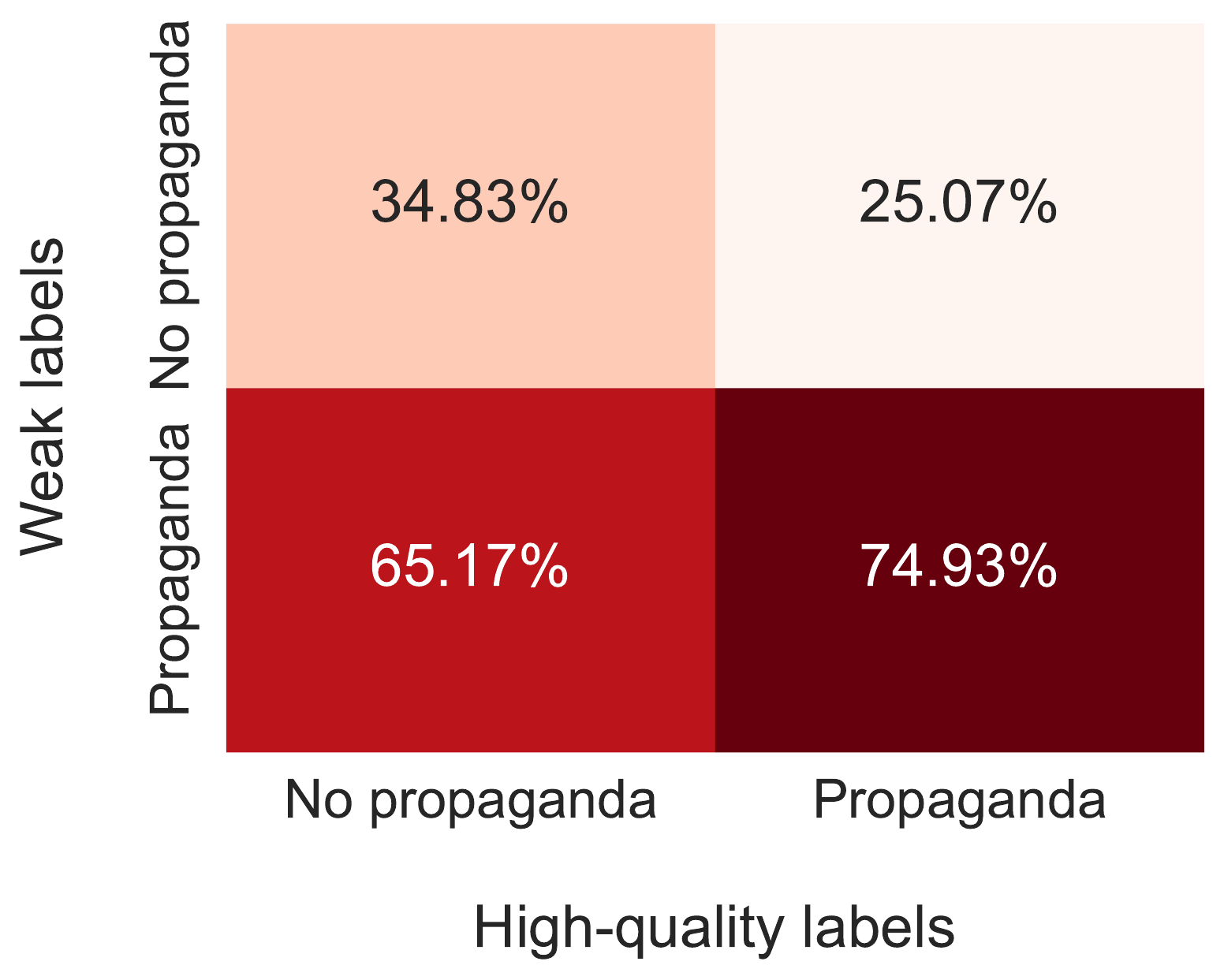}
    \caption{Contingency table comparing weak vs. high-quality labels.}
\label{fig:confusion}
\end{wrapfigure}

\noindent
We find a substantial discrepancy between the weak labels (from \dataset-weak) and our high-quality labels (from human annotation). The overall agreement is only 41.0\% as the majority of labels differ (\Cref{fig:confusion}). Hence, weak labels are noisy and often incorrect, therefore, motivating our use of high-quality labels from human annotation.

\subsection{Propaganda Detection when using Weak vs. High-Quality Labels}


\textbf{RQ2:} \emph{How well can state-of-the-art PLMs detect online propaganda when trained with weak labels vs. high-quality labels?} 

\begin{table*}[h!] 
\centering
\tiny
\renewcommand{\arraystretch}{1.3}
\setlength{\tabcolsep}{2pt}
\begin{tabular}{p{2.3cm}!{\color[rgb]{0.502,0.502,0.502}\vrule}P{0.9cm}P{0.9cm}P{0.9cm}!{\color[rgb]{0.502,0.502,0.502}\vrule}P{0.9cm}P{0.9cm}P{0.9cm}!{\color[rgb]{0.502,0.502,0.502}\vrule}P{0.9cm}P{0.9cm}P{0.9cm}!{\color[rgb]{0.502,0.502,0.502}\vrule}P{0.9cm}P{0.9cm}P{0.9cm}} 
\toprule
                           & \multicolumn{3}{c!{\color[rgb]{0.502,0.502,0.502}\vrule}}{Precision}          & \multicolumn{3}{c!{\color[rgb]{0.502,0.502,0.502}\vrule}}{Recall}        & \multicolumn{3}{c!{\color[rgb]{0.502,0.502,0.502}\vrule}}{F1}        & \multicolumn{3}{c}{AUC}                                              \\
\cmidrule(lr){2-4}\cmidrule(lr){5-7}\cmidrule(lr){8-10}\cmidrule(lr){11-13}
Training data              & BERT                  & RoBERTa               & BERTweet              & BERT                 & RoBERTa              & BERTweet              & BERT                  & RoBERTa              & BERTweet              & BERT                 & RoBERTa               & BERTweet              \\ 
\midrule
TWE \cite{Wang.2020}                       & 14.86 (0.65)          & 14.67 (0.51)           & 14.75 (0.20)          & 46.04 (3.01)        & 45.71 (2.47)         & 53.47 (6.13)          & 22.46 (0.99)          & 22.20 (0.83)         & 23.08 (0.52)          & 47.93 (1.58)         & 48.63 (1.67)          & 47.22 (0.81)          \\
TWEETSPIN \cite{Vijayaraghavan.2022}                       & 23.08 (1.51)          & 23.18 (1.11)          & 23.33 (1.25)          & 60.09 (1.87)         & 59.65 (1.48)         & 59.25 (1.85)          & 33.32 (1.54)          & 33.38 (1.36)         & 33.46 (1.46)          & 64.03 (1.05)         & 63.50 (1.06)          & 63.85 (1.41)          \\ 
\datasetTextOnly-weak (weak labels on our \datasetTextOnly)                       & 16.42 (0.17)           & 16.39 (0.31)           & 16.16 (0.18)          & \textbf{69.24} (1.38) & 69.94 (3.03) & 68.22 (2.64) & 26.55 (0.30)          & 26.56 (0.61)         & 26.13 (0.43)          & 56.71 (0.99)         & 56.79 (2.07)           & 56.64 (0.76)          \\
\midrule
\textbf{\datasetTextOnly~(ours)} & \textbf{61.52} (5.77) & \textbf{66.68} (2.30) & \textbf{68.86} (2.37) & 64.65 (3.83)         & \textbf{70.80} (2.85)          & \textbf{70.65} (2.52)          & \textbf{62.77} (1.92) & \textbf{68.64} (1.80) & \textbf{69.70} (1.31) & \textbf{88.21} (0.62) & \textbf{91.76} (0.62) & \textbf{92.25} (0.80)  \\
\bottomrule
\multicolumn{5}{l}{Stated: mean (SD).}
\end{tabular}
\vspace{-0.2cm}
\caption{Results of propaganda detection for different PLMs fine-tuned on weak vs. high-quality labels.}
\label{tab:performance-weakvshigh}
\end{table*}

\noindent
Table~\ref{tab:performance-weakvshigh} compares the performance of state-of-the-art PLMs in detecting online propaganda when trained with weak labels vs. high-quality labels. For this, we vary the choice of the underlying PLM (BERT, RoBERTa, BERTweet) and the data used for fine-tuning (TWE, TWEETSPIN, \datasetTextOnly). We make the following observations: (1)~The different PLMs reach a similar performance, which corroborates the robustness and reliability of our results. Recall that we intentionally chose state-of-the-art PLMs to allow for comparability when benchmarking the role of weak vs. high-quality labels. (2)~Weak labels from the TWE dataset \cite{Wang.2020} lead to an AUC similar to a random guess, while weak labels from the TWEETSPIN dataset reach an AUC of 64.03. (3)~We use \dataset-weak for a fair comparison where we use the weak labels from our classification into $D_+$ and $D_-$ instead of the high-quality label of \dataset for training. We find an AUC of 56.79. (4)~PLMs trained with high-quality labels perform best with an AUC of 92.25 (for BERTweet). Thereby, we achieve an improvement in AUC over best-performing weak labels (TWEETSPIN) of $\sim$44\%. In sum, the performance gain must be exclusively attributed to the informativeness of high-quality labels (and not other characteristics of the dataset).\footnote{Note that the recall improvement with our high-quality labels is relatively small, while we register a strong improvement in precision. In fact, for weak labels, the fine-tuned models tend to predict the propaganda class too often, which leads to a large number of false positives. In practice, this incurs substantial downstream costs during fact-checking \cite{Naumzik.2022} or may infringe free speech rights.}

\begin{figure}[h!]
\centering
\addtolength{\tabcolsep}{-6.pt}
\begin{tabular}[b]{@{}ccc@{}}
        \includegraphics[width=.5\linewidth]{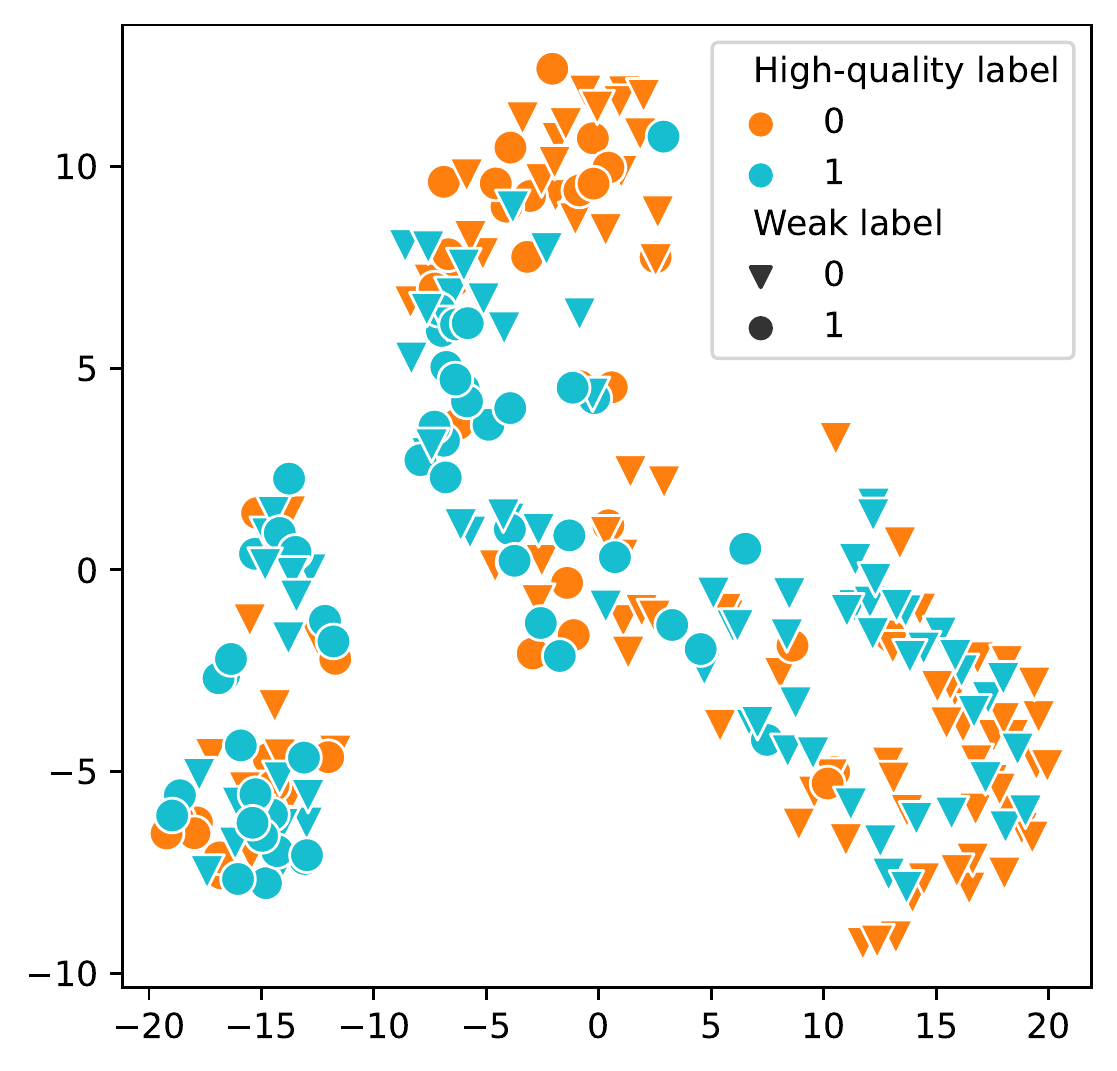}&
        \includegraphics[width=.5\linewidth]{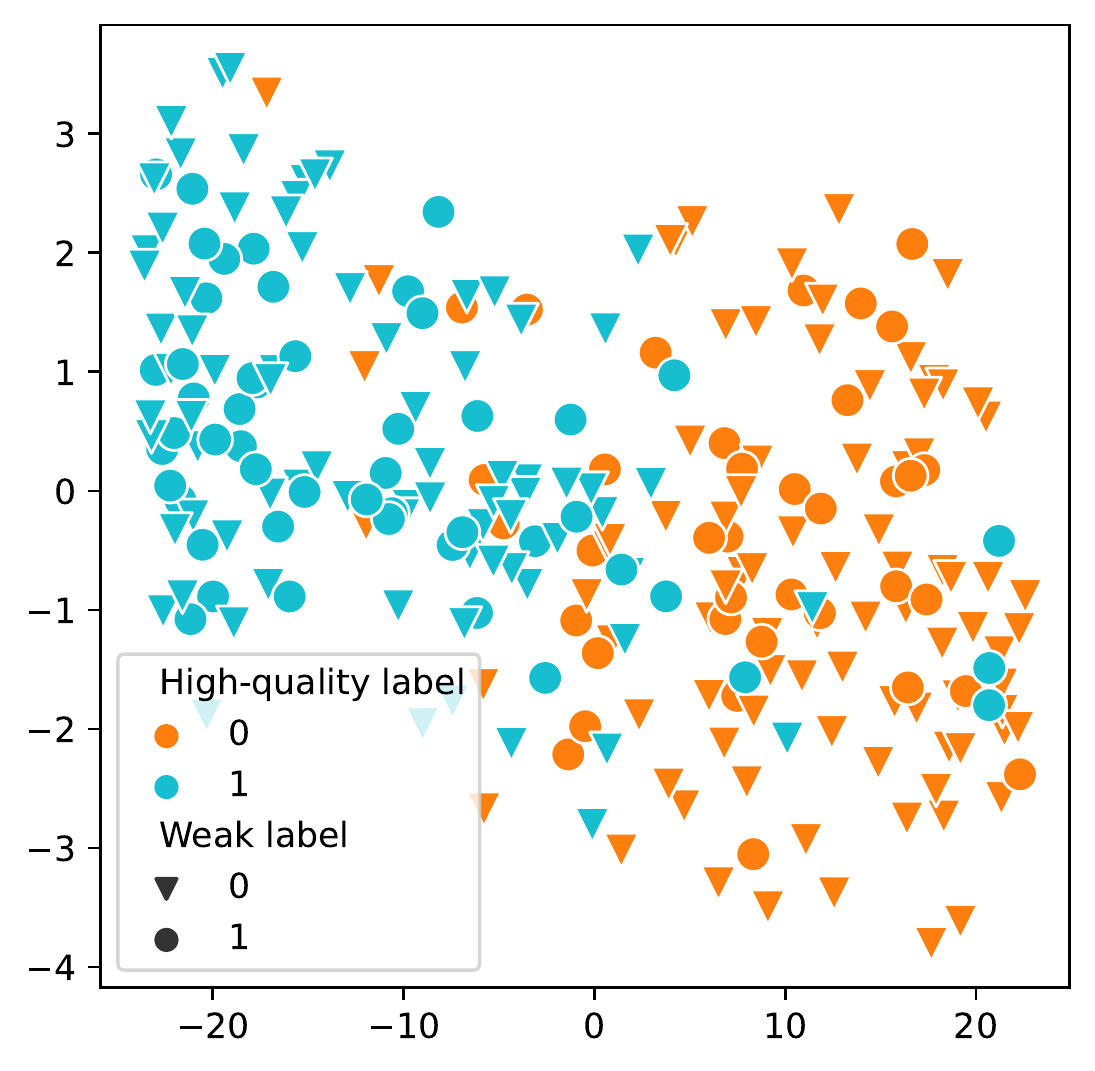} 
\end{tabular}
\vspace{-0.2cm}
\caption{$t$-SNE visualization showing the representations of the $\mathrm{[CLS]}$ tokens for BERTweet fine-tuned on TWEETSPIN labels (left) and \dataset (right).}
\label{fig:tsne-performance}
\vspace{-.5cm}
\end{figure}

We further inspect weak vs. high-quality labels visually. For this, we plot the representation of the $\mathrm{[CLS]}$ tokens from \dataset using $t$-SNE \cite{vandermaaten.2008}. As seen in  \Cref{fig:tsne-performance}, the representations learned with high-quality labels (right plot) are more discriminatory for the true labels than those learned on weak labels (here: TWEETSPIN; left plot).

\subsection{Performance of Few-Shot Learning} \label{sec:exp-fs}


\textbf{RQ3:} \emph{How much can few-shot learning reduce labeling costs for detecting online propaganda?}

\begin{figure}[h!]
	\centering
    \addtolength{\tabcolsep}{-7pt}
    \begin{tabular}[b]{@{}ccc@{}}
        \includegraphics[width=.51\linewidth]{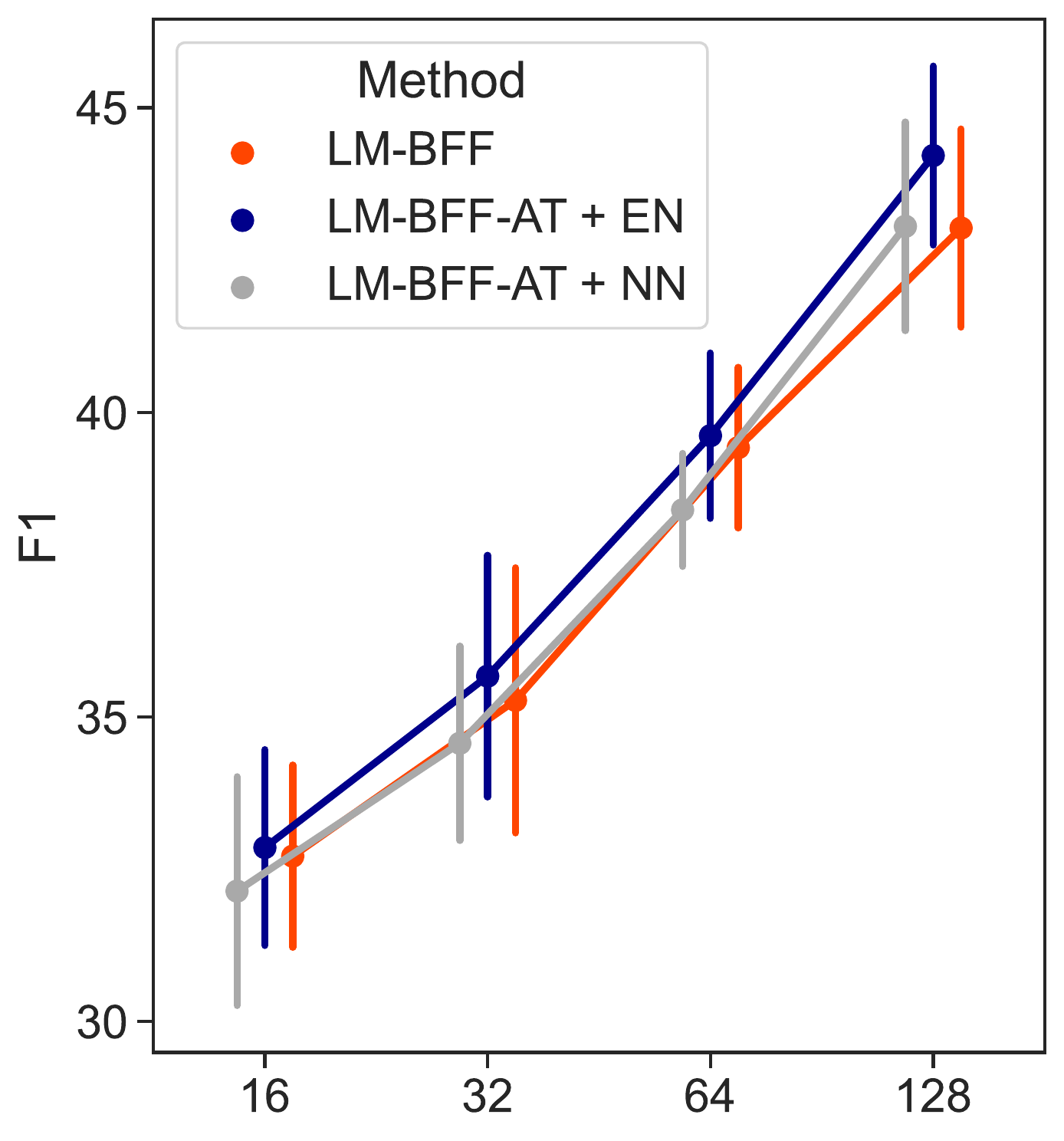}&
        \includegraphics[width=.51\linewidth]{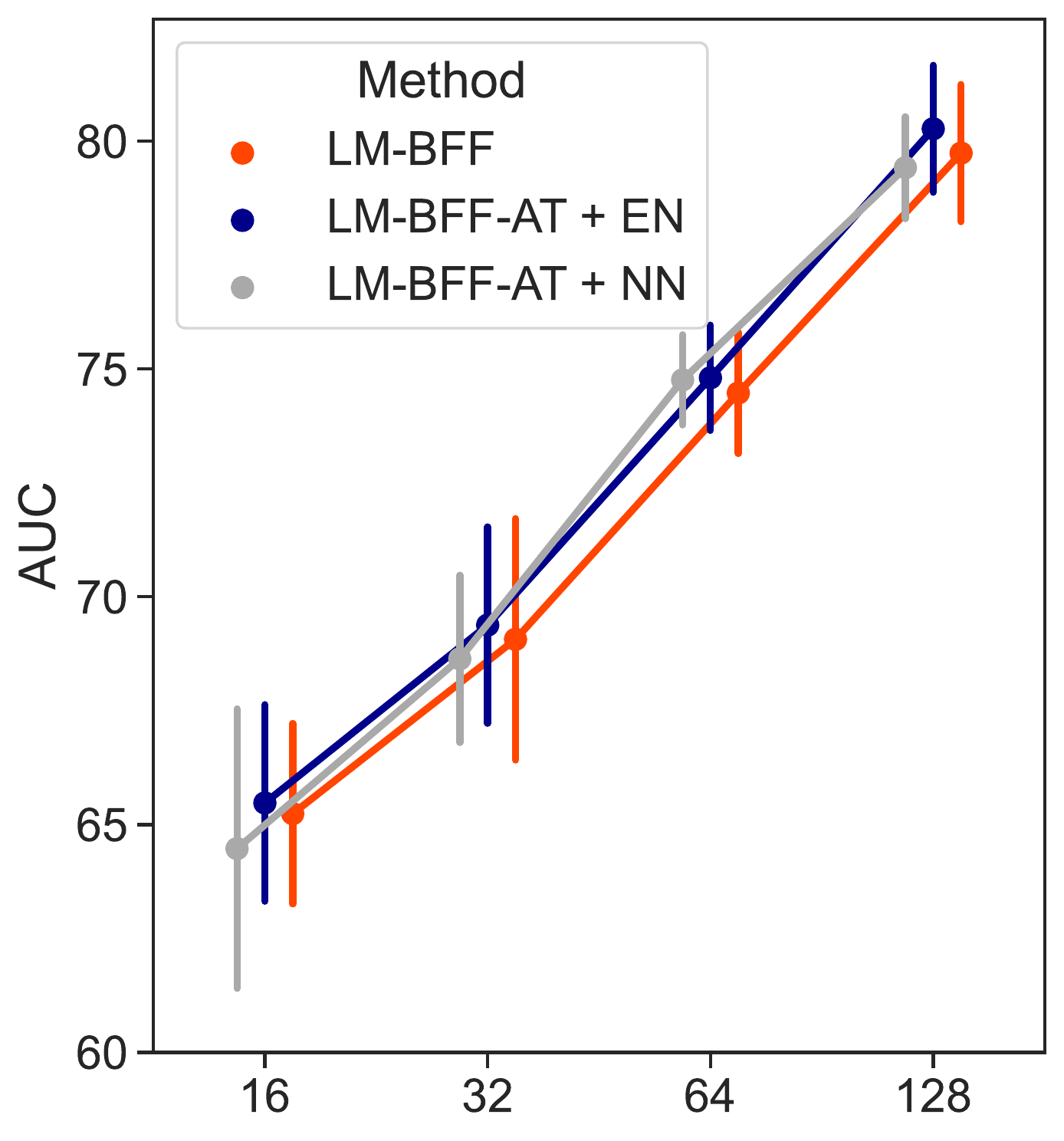} \\
        \includegraphics[width=.51\linewidth]{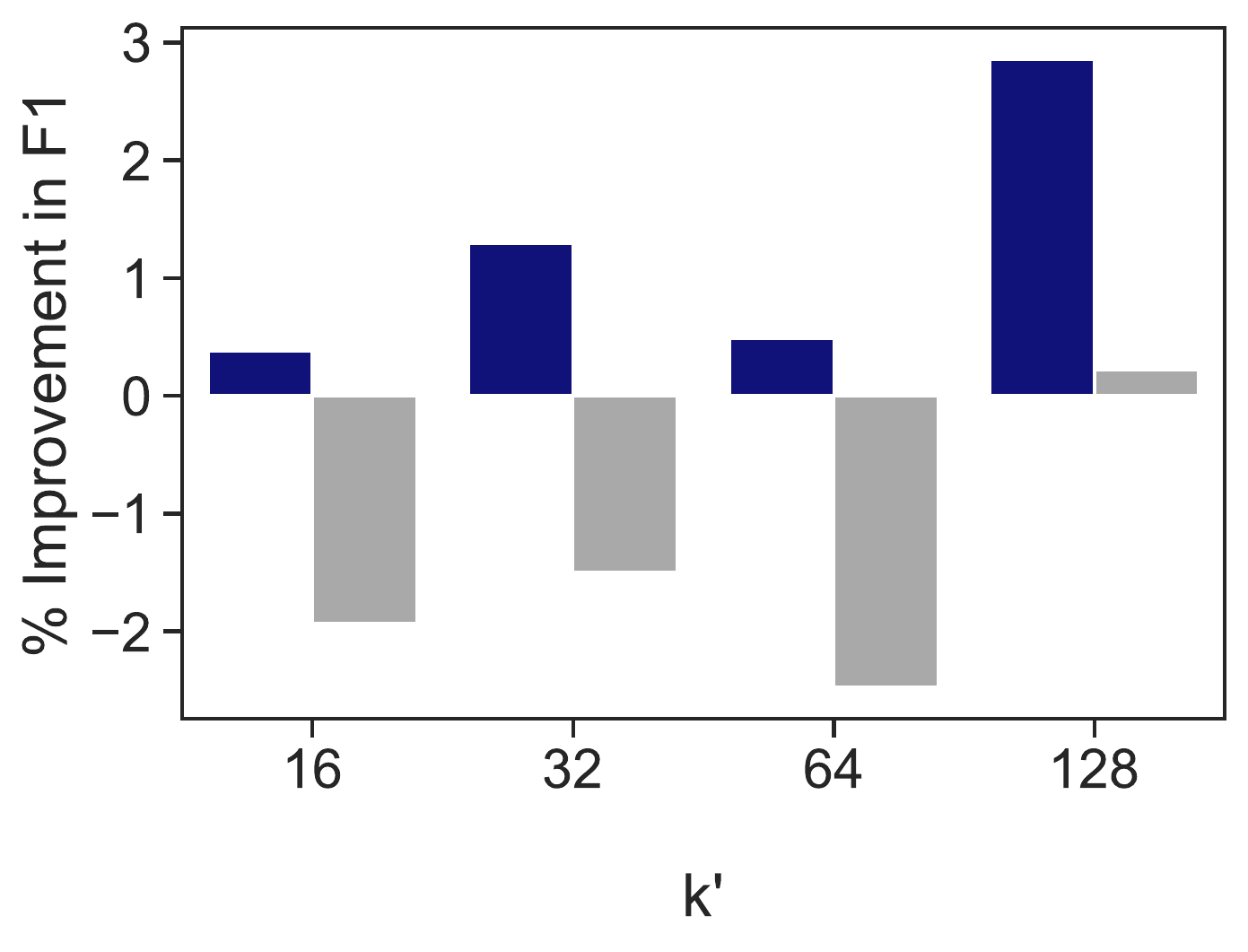}&
        \includegraphics[width=.51\linewidth]{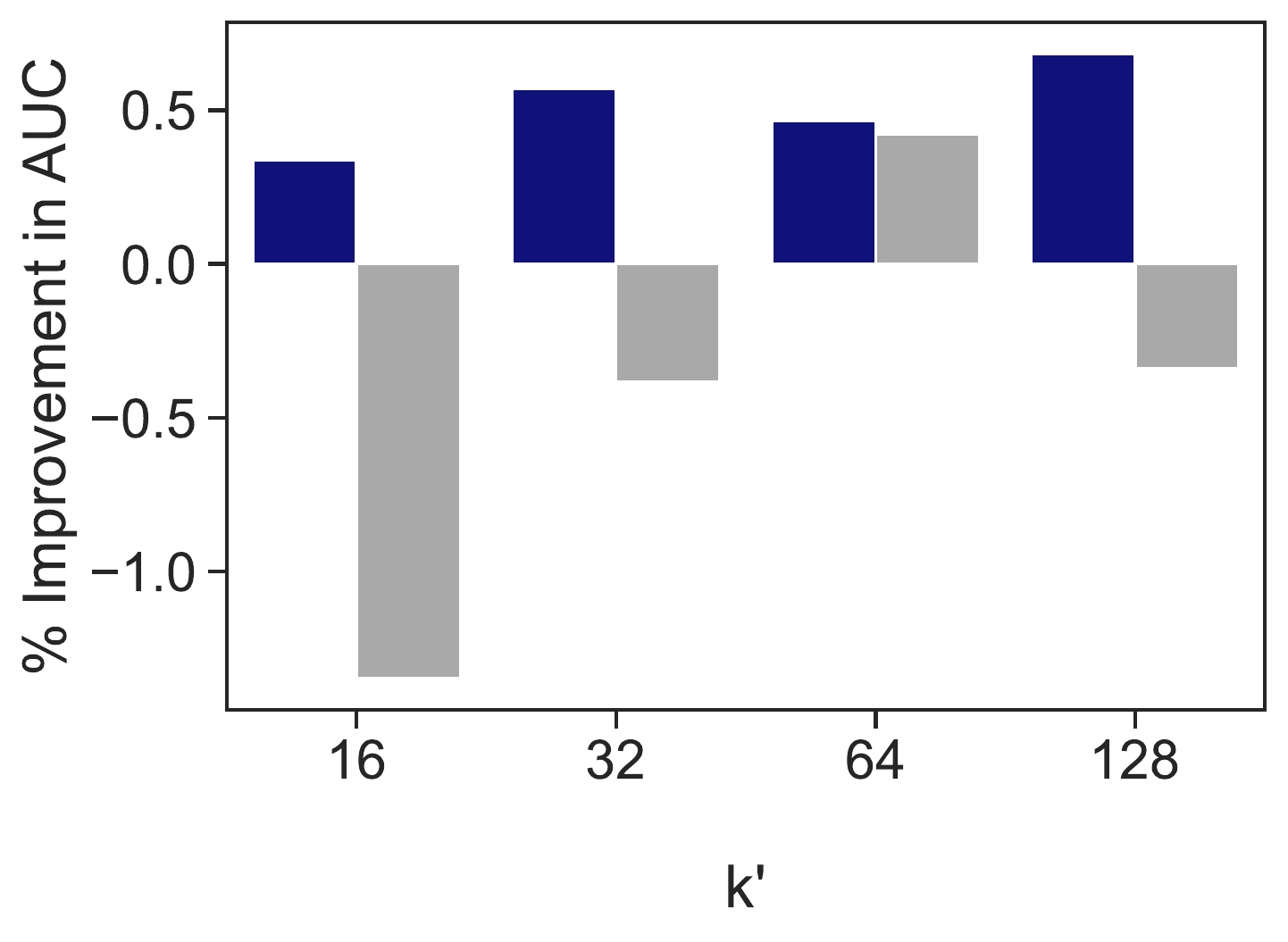}
    \end{tabular}%
    \vspace{-0.5cm}
	\caption{Results for prompt-based learning for LM-BFF vs. LM-BFF-AT (left: F1, right: AUC, top: absolute performance, bottom: \%-improvement over LM-BFF). EN (NN) refers to the elastic net (neural net) classification head. $k'$ refers to the number of examples sampled from each class for both training and validation. Error bars denote the standard errors across 5 runs.}
\label{fig:fs-performance}
\vspace{-0.5cm}
\end{figure}

\noindent
To reduce the costs of labeling, we use few-short learning (see \Cref{fig:fs-performance}). Here, we vary the overall number of labeled samples ($k = 4 \times k'$). We compare the performance of prompt-based learning with LM-BFF (using only \binarylabelshort) vs. LM-BFF-AT (using \binarylabelshort and \strategylabelshort).

\Cref{fig:fs-performance} compares the performance across the different values for $k'$ and the prompt-based learning methods. Generally, a larger $k'$ tends to improve the performance. For example, for $k'=128$ and LM-BFF, we register a mean F1-score of 43.03 and a mean AUC of 79.74. As expected, this is lower than for fine-tuned PLMs but it is a promising finding since only 2.13\% of the labeled examples are used for training and validation. Using LM-BFF-AT with an elastic net as the classification head consistently improves the performance of prompt-based learning across all $k'$. For $k'=128$, we achieve a 2.8\% improvement in the F1-score (44.22) and a 0.7\% improvement in AUC (80.27). On average, over all $k'$, the improvement amounts to 1.25\% for the F1-score and 0.51\% for the AUC. Generally, the variant with an elastic net tends to be better than the variant with a neural network, likely due to the small size of the training sample.\footnote{We  report evaluations of the auxiliary task (i.e., prompt-based learning for \strategylabelshort) in \Cref{sec:at-performance}. We evaluate increasing the number of few-shot samples $k'$ for our prompt-based learning in \Cref{sec:exp-fs-higherk}.}


\subsection{Propaganda Detection across Different Contexts} \label{sec:exp-topicshifts}

\textbf{RQ4:} \emph{What is the cross-context performance in detecting propaganda?}

\begin{table*}[h!]
\centering
\tiny
\renewcommand{\arraystretch}{1.3}
\setlength{\tabcolsep}{2pt}
\begin{tabular}{p{0.8cm}!{\color[rgb]{0.502,0.502,0.502}\vrule}P{0.9cm}P{0.9cm}P{0.9cm}!{\color[rgb]{0.502,0.502,0.502}\vrule}P{0.9cm}P{0.9cm}P{0.9cm}!{\color[rgb]{0.502,0.502,0.502}\vrule}P{0.9cm}P{0.9cm}P{0.9cm}!{\color[rgb]{0.502,0.502,0.502}\vrule}P{0.9cm}P{0.9cm}P{0.9cm}} 
\toprule
                           & \multicolumn{3}{c!{\color[rgb]{0.502,0.502,0.502}\vrule}}{Precision}          & \multicolumn{3}{c!{\color[rgb]{0.502,0.502,0.502}\vrule}}{Recall}        & \multicolumn{3}{c!{\color[rgb]{0.502,0.502,0.502}\vrule}}{F1}        & \multicolumn{3}{c}{AUC}                                              \\
\cmidrule(lr){2-4}\cmidrule(lr){5-7}\cmidrule(lr){8-10}\cmidrule(lr){11-13}
Context &          BERT &        RoBERTa &      BERTweet &           BERT &       RoBERTa &      BERTweet &          BERT &       RoBERTa &      BERTweet &          BERT &       RoBERTa &      BERTweet \\
\midrule
\multirow{2}{*}{\protect\circledblue{R}}   &  63.02 (2.89) &   66.03 (2.58) &  65.75 (4.79) &   61.61 (2.85) &    68.60 (0.9) &  70.42 (2.97) &  62.21 (1.03) &  67.26 (1.06) &  67.83 (1.37) &  87.31 (0.65) &  91.48 (0.69) &  91.77 (0.51) \\
\multirow{2}{*}{\protect\circledorange{M}} &  55.31 (8.93) &  62.18 (13.54) &  62.72 (6.64) &  61.16 (17.69) &  53.9 (17.87) &  48.83 (9.18) &  56.25 (5.62) &  54.23 (9.15) &  54.07 (4.02) &  74.07 (5.83) &  75.72 (3.56) &  78.05 (1.79) \\
\multirow{2}{*}{\protect\circledgreen{U}}   &  69.55 (7.81) &   76.24 (7.16) &  78.67 (5.37) &   55.99 (5.65) &  51.37 (5.51) &  56.51 (9.51) &  61.76 (4.62) &  61.13 (4.68) &  65.26 (6.55) &  78.12 (3.93) &  79.96 (2.76) &   82.47 (3.1) \\
\arrayrulecolor{black!30}\midrule
\multirow{2}{*}{Overall}  &   62.56 (3.0) &    65.8 (2.23) &  65.94 (4.44) &   61.29 (3.32) &  66.78 (1.28) &  68.25 (3.31) &  61.81 (1.19) &  66.27 (1.35) &   66.9 (0.93) &  86.86 (0.49) &   90.67 (0.9) &  91.15 (0.47) \\
\arrayrulecolor{black}\bottomrule
\multicolumn{5}{l}{Stated: mean (SD).}
\end{tabular}
\vspace{-0.2cm}
\caption{Results of propaganda detection for different PLMs fine-tuned on \datasetplus. We show the performance of PLMs in detecting propaganda across three contexts (i.e., \protect\circledblue{R} Russian propaganda, \protect\circledorange{M} anti-Muslim propaganda in India, and \protect\circledgreen{U} anti-Uyghur propaganda).}
\label{tab:performance-crosscontexts}
\end{table*}

\noindent
\Cref{tab:performance-crosscontexts} compares the performance of PLMs for detecting propaganda across contexts. We fine-tune the PLMs on \datasetplus, which includes the contexts of \protect\circledblue{R} Russian propaganda ($N=\num{30000}$), \protect\circledorange{M} anti-Muslim propaganda in India ($N=\num{1000}$), and \protect\circledgreen{U} anti-Uyghur propaganda ($N=\num{1000}$). Overall, we find that the performance is largely robust across contexts. The performance of detecting propaganda in the two additional contexts is considerably high. This is especially surprising when keeping in mind that both events individually only account for around $\sim$3\% of the entire \datasetplus dataset.

\section{Discussion}

We introduce \dataset: the first large-scale dataset for online propaganda detection with human annotations. Our experiments have direct implications for the NLP community.

\noindent
$\bullet$\,\textbf{Implication 1:} \emph{When identifying propaganda, there is a substantial discrepancy between weak labeling and human annotations.} This pinpoints weaknesses in existing datasets for online propaganda detection \cite{Wang.2020, Vijayaraghavan.2022}, since these make exclusive use of weak labeling. To this end, our work highlights the importance of human feedback for sensitive NLP tasks such as propaganda detection. 

\noindent
$\bullet$\,\textbf{Implication 2:} \emph{High-quality labels are crucial to detect online propaganda.} Our experiments are intentionally based on state-of-the-art PLMs to ensure the reliability and comparability of our results. Generally, PLMs fail to detect propaganda when fine-tuned with weak labels. In contrast, there is a large improvement ($\sim$44\%) when using high-quality labels. 

\noindent
$\bullet$\,\textbf{Implication 3:} \emph{Few-shot learning can be an effective remedy to reduce the cost of human annotation of propaganda.} To the best of our knowledge, our work is the first to adapt few-shot learning (via prompt-based learning) to propaganda detection. Interestingly, our performance is similar to that in related NLP tasks such as, e.g., detecting rumors \cite{Lin.2023} and humor \cite{Li.2023}. Despite the challenging nature of our task, the performance of few-shot learning is promising. For example, only $k = 64$ ($k'=16$) high-quality annotated samples are needed to outperform propaganda detection with weak labels. For $k = 512$ ($k'=128$), we already achieve an improvement over weak labels of 24.54\%.

\section{Limitations}

As with other works, ours is not free of limitation. First, there is no universal rule to identify propaganda. Hence, the perception may vary across individuals. We address this by having our dataset annotated through multiple raters and showing raters a task description that includes a widely accepted definition of propaganda \cite[see][]{Smith.2022}. In addition, we carefully selected our inclusion criteria for data collection to ensure a broad and diverse set of posts. Second, we are further aware that PLMs may embed biases that are populated in downstream tasks. Hence, we call for careful use when deploying our methods in practice. Third, narratives that fall under the scope of propaganda may change over time. Hence, we recommend that both the dataset construction and the PLM fine-tuning is repeated regularly. To this end, we provide a cost-effective approach through few-shot learning.  

\section{Ethics Statement}

Our dataset will benefit research on improving social media integrity. The construction of our dataset follows best-practice for ethical research \cite{Rivers.2014}. The dataset construction and usage were approved as ethically unproblematic by the ethics commission of the Faculty of Mathematics, Informatics and Statistics at LMU Munich (ethics approval number: EK-MIS-2023-160). In particular, our dataset contains only publicly available information. The privacy policy of Twitter/X warns users that their content can be viewed by the general public. Further, we respect the privacy of users and only report aggregate results throughout our paper. Although we believe the intended use of this work is largely positive, there exists potential for misuse (e.g., by propaganda campaigns to run adversarial attacks and develop techniques to avoid detection). To this end, we call for meaningful research by the NLP community to further improve social media integrity. Finally, we encourage careful use of our dataset, as it contains potentially offensive language and manipulative content, which lies in the nature of the task. 

\dataset (and \datasetplus) was annotated using Prolific. The workers were paid 11.40 USD per hour, which is above the federal minimum wage.

\section{Acknowledgments}

We would like to thank the anonymous reviewers for their helpful comments. Funding by the German Research Foundation (Grant: 543018872) is acknowledged.

\bibliography{anthology,custom}

\clearpage
\appendix
\counterwithin{figure}{section}
\counterwithin{table}{section}

\sloppy
\raggedbottom

\section{Dataset Construction}

\subsection{\datasetplus}\label{sec:hqp+}

\datasetplus is an extended dataset where we include broad coverage of different propaganda themes. To this end, we additionally include (1)~\protect\circledorange{M}~anti-Muslim propaganda in India and (2)~\protect\circledgreen{U}~anti-Uyghur propaganda. Our choice is informed by prior social media research \cite{Oxford.2023}. Both events have diverse origins and are considered as salient propaganda topics in current media studies \cite{Oxford.2023}. 

For both \protect\circledorange{M} and \protect\circledgreen{U} propaganda, we collected and annotated $\num{1000}$ posts. Hence, the final dataset comprises overall  $N_+=\num{32000}$ posts. Our aim is to provide a dataset based on which PLMs. can be fine-tuned to detect general patterns of propaganda

Our annotation follows the same procedure as in \Cref{sec:data_construction}: based on boosted sampling and human annotation (i.e., multi-annotator and multi-batch approach) we generate high-quality labels. As the topics differ, we use different keywords to obtain candidate posts for each class of the two new events. We refer to \Cref{sec:keywords} for a full list of keywords.

\subsection{Keywords for Dataset Construction}\label{sec:keywords}

\Cref{tab:accusation_phrases_R} (\protect\circledblue{R}), \Cref{tab:accusation_phrases_H} (\protect\circledorange{M}), and \Cref{tab:accusation_phrases_U} (\protect\circledgreen{U}) list the keywords used in our dataset construction process to obtain candidate posts for the positive class ($D_+$) via accusations in replies. \Cref{tab:warrelated_phrases_R} (\protect\circledblue{R}), \Cref{tab:warrelated_phrases_H} (\protect\circledorange{M}), and \Cref{tab:warrelated_phrases_U} (\protect\circledgreen{U}) show the keywords that are used to collect candidate posts for the negative class ($D_-$). Generally, keywords relevant to the positive class should mostly be terms that express accusations of propaganda, while keywords relevant to the negative class should be mostly terms that refer to general activities of the war.  The keywords in both lists contain further results from our construction procedure in that we list the iteration in which they were added to the list. 
\vspace{-0.2cm}

\begin{table}[H]
	\centering
	\footnotesize
	\singlespacing
        \setlength{\tabcolsep}{6pt}
    \begin{tabular}{lc}
        \toprule
        Keywords ($D_+$) & Iteration\\
        \midrule
        \texttt{russia(n) $\land$ propaganda} & 1 \\
        \texttt{russia(n) $\land$ propagandist} & 1 \\
        \texttt{kremlin $\land$ propaganda} & 2 \\
        \texttt{kremlin $\land$ propagandist} & 2 \\
        \texttt{putinist(s)} & 2 \\
        \texttt{putinism} & 2 \\
        \texttt{russia(n) $\land$ lie(s)} & 3 \\
        \texttt{war $\land$ propaganda} & 3 \\
        \texttt{war $\land$ lie(s)} & 3  \\
        \texttt{putin $\land$ propaganda} & 3 \\
        \texttt{putin $\land$ propagandist} & 3 \\
        \texttt{russia(n) $\land$ fake news} & 3 \\
        \bottomrule
    \end{tabular}
    \caption{\protect\circledblue{R} List of keywords used to get propaganda accusations in the context of Russian propaganda and the corresponding iteration they were added to the keyword list. The $\land$-operator indicates that both keywords have to appear.}
    \label{tab:accusation_phrases_R}
\end{table}
\vspace{-0.7cm}

\begin{table}[H]
	\centering
	\footnotesize
	\singlespacing
        \setlength{\tabcolsep}{6pt}
    \begin{tabular}{lc}
        \toprule
        Keywords ($D_+$) & Iteration\\
        \midrule
        \texttt{anti(-)muslim $\land$ propaganda} & 1 \\
        \texttt{anti(-)muslim $\land$ hate} & 1 \\
        \texttt{india(n) $\land$ propagandist} & 2 \\
        \texttt{india(n) $\land$ propaganda} & 2 \\
        \texttt{india(n) $\land$ lie(s)} & 3 \\
        \texttt{hindutva $\land$ propaganda} & 3 \\
        \texttt{hindutva $\land$ lie(s)} & 3 \\
        \texttt{hindutva $\land$ hate} & 3 \\
        \texttt{hindutva $\land$ conspiracy} & 3 \\
        \bottomrule
    \end{tabular}
    \caption{\protect\circledorange{M} List of keywords used to get propaganda accusations in the context of anti-Muslim propaganda in India and the corresponding iteration they were added to the keyword list. The $\land$-operator indicates that both keywords have to appear.}
    \label{tab:accusation_phrases_H}
\end{table}
\vspace{-0.7cm}

\begin{table}[H]
	\centering
	\footnotesize
	\singlespacing
        \setlength{\tabcolsep}{6pt}
    \begin{tabular}{lc}
        \toprule
        Keywords ($D_+$) & Iteration\\
        \midrule
        \texttt{anti(-)uyghur $\land$ propaganda} & 1 \\
        \texttt{anti(-)uyghur $\land$ hate} & 1 \\
        \texttt{covid19 $\land$ propaganda} & 1 \\
        \texttt{covid19 $\land$ conspiracy} & 1 \\
        \texttt{china $\land$ propagandist} & 2 \\
        \texttt{china $\land$ propaganda} & 2 \\
        \texttt{chinese $\land$ propagandist} & 2 \\
        \texttt{chinese $\land$ propaganda} & 2 \\
        \texttt{china $\land$ lie(s)} & 3 \\
        \texttt{chinese $\land$ lie(s)} & 3 \\
        \texttt{beijing $\land$ propaganda} & 3 \\
        \texttt{beijing $\land$ lie(s)} & 3 \\
        \texttt{beijing $\land$ hate} & 3 \\
        \texttt{beijing $\land$ conspiracy} & 3 \\
        \bottomrule
    \end{tabular}
    \caption{\protect\circledgreen{U} List of keywords used to get propaganda accusations in the context of anti-Uyghur propaganda and the corresponding iteration they were added to the keyword list. The $\land$-operator indicates that both keywords have to appear.}
    \label{tab:accusation_phrases_U}
\end{table}
\vspace{-0.7cm}

\begin{table}[H]
	\centering
	\footnotesize
	\singlespacing
    \begin{tabular}{lc}
        \toprule
        Keywords ($D_-$) & Iteration\\
        \midrule
        \texttt{russia $\land$ war} & 1 \\
        \texttt{ukraine $\land$ war} & 1 \\
        \texttt{\#istandwithrussia} & 1 \\
        \texttt{\#istandwithputin} & 1 \\
        \texttt{russian $\land$ war} & 2 \\
        \texttt{ukrainian $\land$ war} & 2 \\
        \texttt{\#russianukrainianwar} & 2 \\
        \texttt{\#ukrainerussiawar} & 2 \\
        \texttt{\#standwithrussia} & 2 \\
        \texttt{\#standwithputin} & 2 \\
        \texttt{\#russia} & 2 \\
        \texttt{\#russiaukraine} & 2 \\
        \texttt{\#ukraine} & 2 \\
        \texttt{putin $\land$ war} & 3 \\
        \texttt{\#putin} & 3 \\
        \texttt{\#lavrov} & 3 \\
        \texttt{\#zakharova} & 3 \\
        \texttt{\#nato} & 3 \\
        \texttt{\#donbass} & 3 \\
        \texttt{\#mariupol} & 3 \\
        \bottomrule
    \end{tabular}
    \caption{\protect\circledblue{R} List of keywords used to get war-related posts in the context of Russian propaganda and the corresponding iteration they were added to the keyword list. The $\land$-operator indicates that both keywords have to appear.}
    \label{tab:warrelated_phrases_R}
    \vspace{-0.5cm}
\end{table}

\begin{table}[H]
	\centering
	\footnotesize
	\singlespacing
    \begin{tabular}{lc}
        \toprule
        Keywords ($D_-$) & Iteration\\
        \midrule
        \texttt{india $\land$ bulli} & 1 \\
        \texttt{india $\land$ sulli} & 1 \\
        \texttt{\#lovejihaad} & 1 \\
        \texttt{\#coronajihaad} & 1 \\
        \texttt{muslim $\land$ india} & 2 \\
        \texttt{islam $\land$ india} & 2 \\
        \texttt{\#lovejihad} & 2 \\
        \texttt{\#coronajihad} & 2 \\
        \texttt{\#coronaterrorism} & 2 \\
        \texttt{\#coronabombstablighi} & 2 \\
        \texttt{\#islamindia} & 3 \\
        \texttt{\#muslimindia} & 3 \\
        \texttt{\#romeojihaad} & 3 \\
        \texttt{\#romeojihad} & 3 \\
        \bottomrule
    \end{tabular}
    \caption{\protect\circledorange{M} List of keywords used to get related posts in the context of anti-Muslim propaganda in India and the corresponding iteration they were added to the keyword list. The $\land$-operator indicates that both keywords have to appear.}
    \label{tab:warrelated_phrases_H}
    \vspace{-0.5cm}
\end{table}

\begin{table}[H]
	\centering
	\footnotesize
	\singlespacing
    \begin{tabular}{lc}
        \toprule
        Keywords ($D_-$) & Iteration\\
        \midrule
        \texttt{uyghur $\land$ terrorist(s)} & 1 \\
        \texttt{\#xinjiang} & 1 \\
        \texttt{\#forcedlabor} & 1 \\
        \texttt{xinjiang $\land$ forced $\land$ labor} & 2 \\
        \texttt{china $\land$ forced $\land$ labor} & 2 \\
        \texttt{\#pompeo} & 2 \\
        \texttt{\#genocide} & 2 \\
        \texttt{\#uyghur} & 2 \\
        \texttt{pompeo $\land$ forced $\land$ labor} & 3 \\
        \texttt{\#uyghurgenocide} & 3 \\
        \texttt{\#usvirus} & 3 \\
        \bottomrule
    \end{tabular}
    \caption{\protect\circledgreen{U} List of keywords used to get related posts in the context of anti-Uyghur propaganda and the corresponding iteration they were added to the keyword list. The $\land$-operator indicates that both keywords have to appear.}
    \label{tab:warrelated_phrases_U}
    \vspace{-0.5cm}
\end{table}

\section{Annotation}\label{appendix:annotation_instructions}
\subsection{Construction of Annotation Guidelines}

\begin{figure*}[p]
    \captionsetup{position=top}
    \centering
    \includegraphics[width=0.95\linewidth]{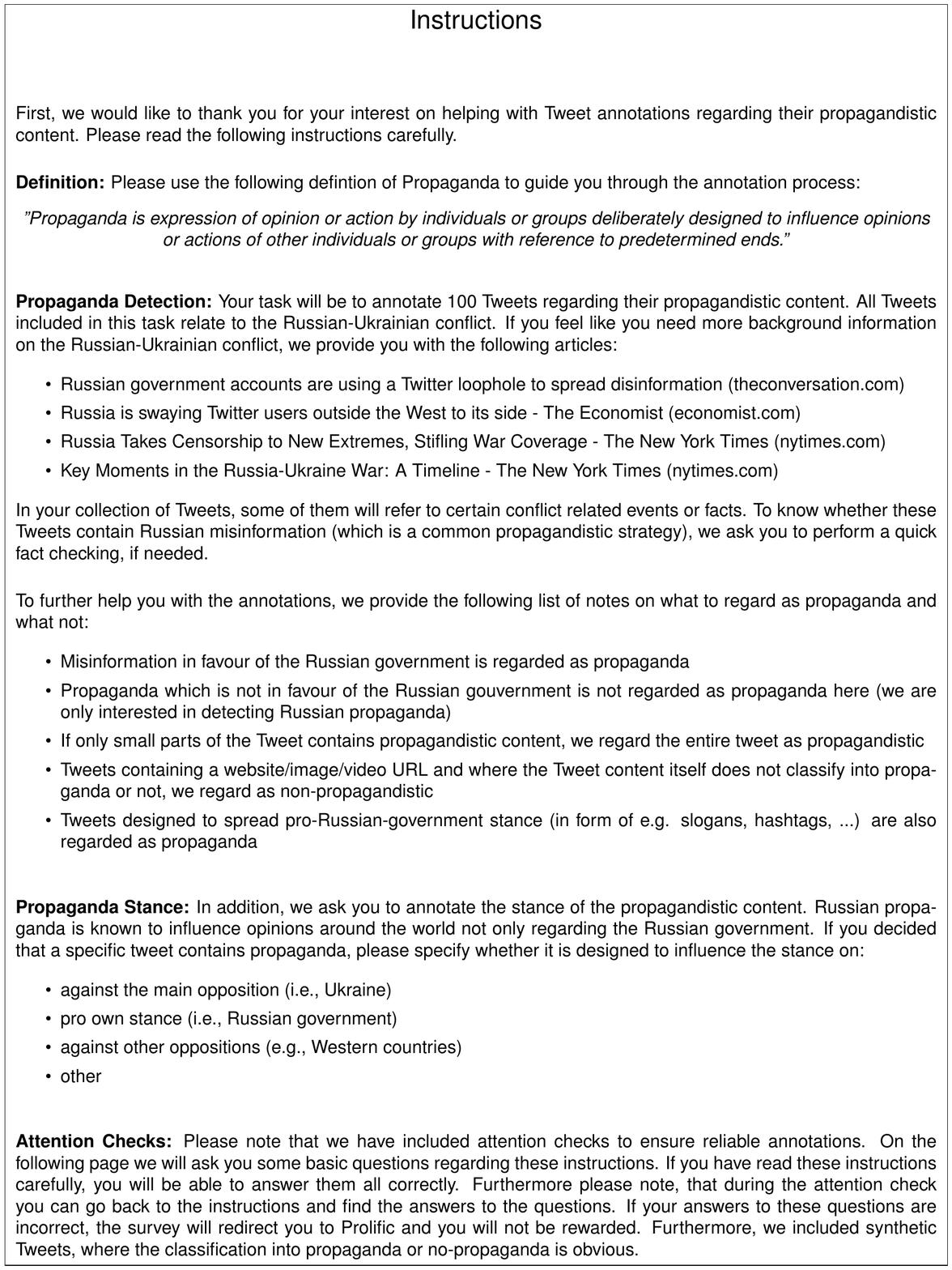}
    \caption{\protect\circledblue{R} Instructions for annotators for the context of Russian propaganda.}
\label{fig:instructions_r}
\end{figure*}

\begin{figure*}[p]
    \captionsetup{position=top}
    \centering
    \includegraphics[width=0.95\linewidth]{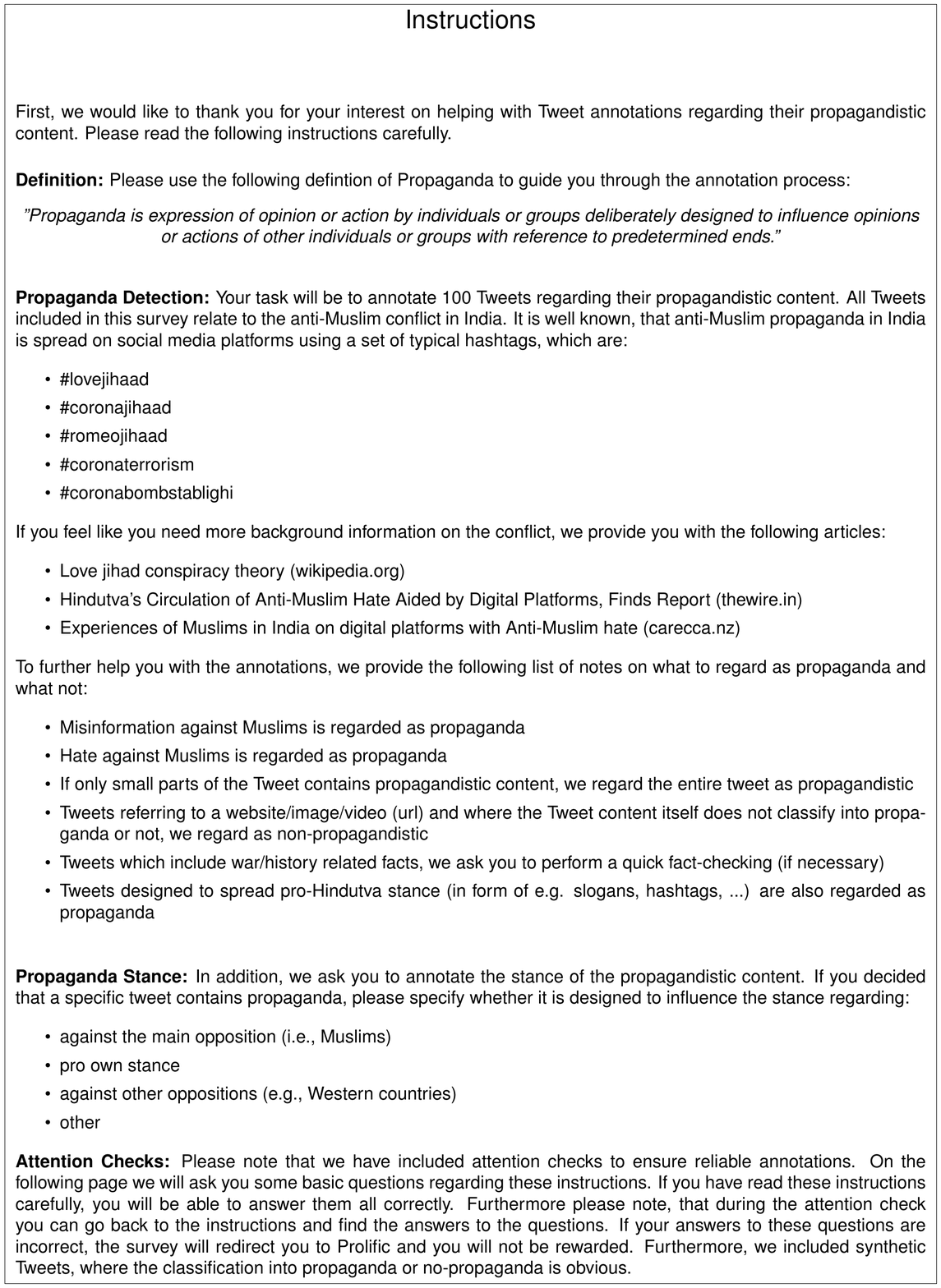}
    \caption{\protect\circledorange{M} Instructions for annotators for the context of anti-Muslim propaganda in India.}
\label{fig:instructions_i}
\end{figure*}

\begin{figure*}[p]
    \captionsetup{position=top}
    \centering
    \includegraphics[width=0.95\linewidth]{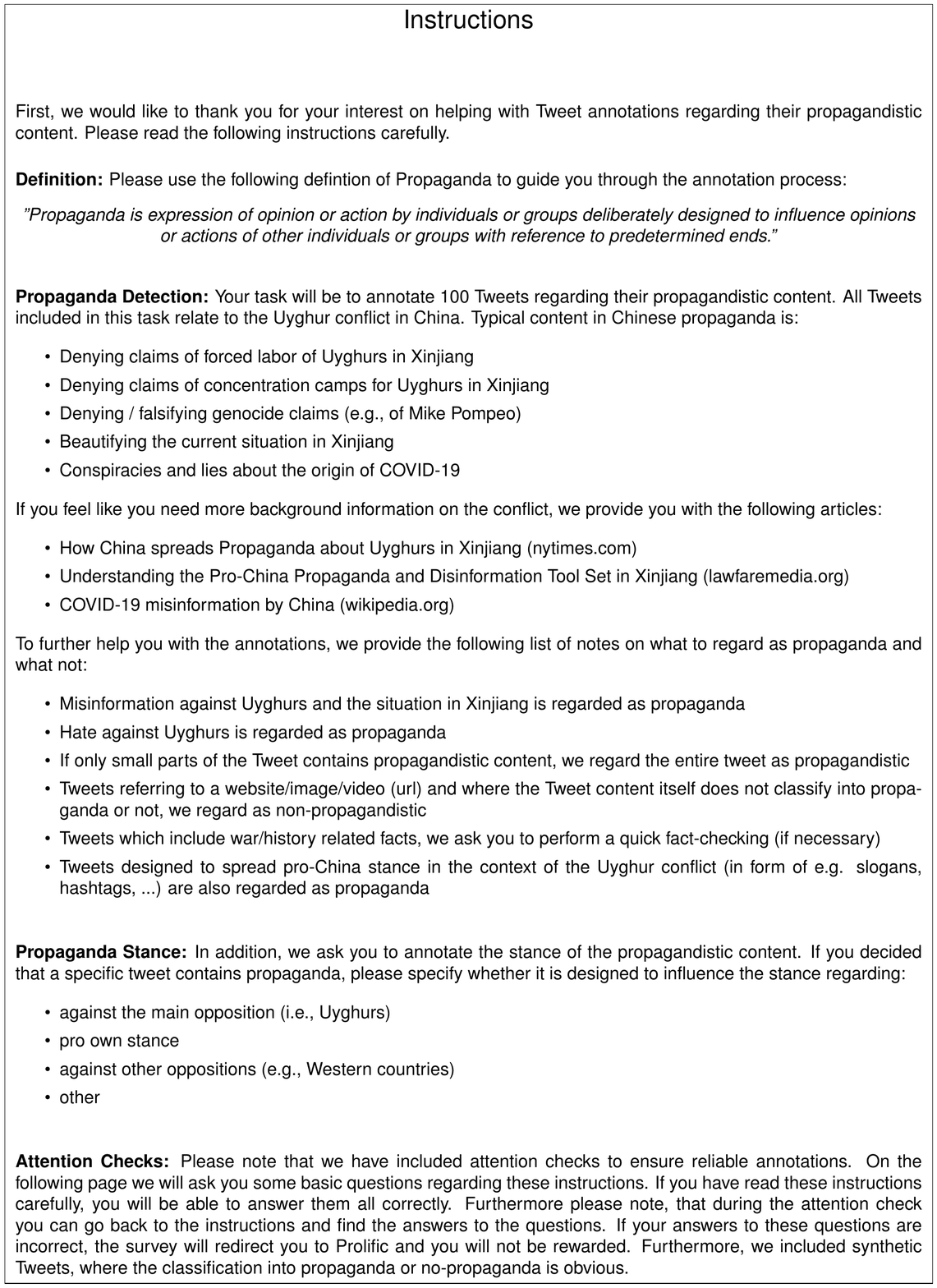}
    \caption{\protect\circledgreen{U} Instructions for annotators for the context of anti-Uyghur propaganda.}
\label{fig:instructions_c}
\end{figure*}

\Cref{fig:instructions_r} (\protect\circledblue{R}), \Cref{fig:instructions_i} (\protect\circledorange{M}), and \Cref{fig:instructions_c} (\protect\circledgreen{U}) show the instructions of batch annotations we present to the workers on Prolific (\url{https://www.prolific.co/}). We follow best practices \cite{Song.2020}. That is, we provide a detailed and comprehensible description of the task, a precise definition of the labels, and a transparent disclosure that we use attention checks. 

\subsection{Data collected from Workers}

We asked all workers for their agreement on using data collected during the study. We did not collect, use, or publish any sensitive data (i.e., racial, ethnic origin, religious or political beliefs, or health status). We only collected the Prolific-id for payment, their agreement on the data collection and usage, and their annotations. By default, public account information (name, e-mail, etc.) is stored in the output files, which we however neither use nor publish in any form. We only used their answers to the study (i.e., their annotations to the Tweets) together with the answers of other study participants to determine the labels of each tweet. We do not publish individual annotations.

\subsection{Diversity in Annotations}

It is crucial to consider the potential impact of worker diversity on the labeling process, especially in tasks like propaganda detection. During our annotation, it was necessary to guarantee that the selection of workers was fluent in English. Thereby, we ensured an accurate understanding and interpretation of the social media content, which was exclusively in English. We took several measures to mitigate bias. These included providing detailed guidelines and background information on each propaganda context and employing a multi-annotator and multi-batch approach to balance out individual biases. Additionally, strict attention checks and validation steps were implemented to maintain annotation quality. Lastly, we average over several workers (and thus different worker backgrounds), but we oversee only little variability across worker backgrounds (Cohen’s Kappa of $0.71$).

For our quality checks, we used internal annotations. Specifically, for each batch of 100 posts, we manually annotate 10 posts with respect \binarylabelshort to measure the quality of the annotations. Hence, the labels of internally annotated tweets only affect the decision of whether the batch annotation has to be repeated. Therefore, our labels do not determine any final label in \dataset and \datasetplus. Internal annotations for the quality checks might limit the diversity of interpretations, especially in the presence of a bias towards a particular understanding of propaganda. To address this, we ensured that the team of propaganda experts responsible for internal annotations is aware of the broad range of perspectives. The team is well-versed in the literature on propaganda and the three contexts included in \datasetplus. The awareness and expertise were crucial for minimizing potential biases due to subjectivity.

\subsection{Examples of Disagreement}

\Cref{tab:exampleposts_disagreement} lists five example posts where the annotators initially disagreed on \binarylabelshort. Disagreements are a natural aspect of human annotation, especially in subjective tasks such as propaganda detection. That is why we rely on the following strategy to resolve the disagreements: the posts with disagreement are re-annotated by randomly assigning them to one of the top 25 annotators. The top 25 annotators are selected based on the inter-annotator agreement and the percentage of correctly annotated tweets of our internal annotation.

\begin{table}[h!]
\centering
\scriptsize
\renewcommand{\arraystretch}{1.3}
\begin{tabular}{p{7cm}}
\toprule
Posts                                                                                                                                                                                                                                                                                             \\
\midrule
``\texttt{@USER @USER Lol be civil, but also vote communism}''      \\
``\texttt{If you still support Biden, you owe America an apology.}''       \\
``\texttt{Putin Says Russia ‘Has Not Lost Anything’}''      \\
``\texttt{Putin’s War Could Save the Global Economic Order HTTPURL}''            \\
``\texttt{@USER @USER it's not Ukraine's job to make everyone happy.}'' \\
\bottomrule
\end{tabular}
\caption{Examples where the annotators initially disagreed on \binarylabelshort for \dataset.}
\label{tab:exampleposts_disagreement}
\end{table}
\vspace{-0.6cm}

\subsection{Examples of false Annotations}

As with other works on datasets, ours is not free of false annotations. \Cref{tab:exampleposts_false} lists three example posts of false annotations.

\begin{table}[h!]
\centering
\scriptsize
\renewcommand{\arraystretch}{1.3}
\begin{tabular}{p{6cm}P{0.5cm}}
\toprule
Posts      & \binarylabelshort                                                                                                                                                                                                                                                                                       \\
\midrule
``\texttt{@USER @USER @USER It appears you have no idea about how the US propaganda works.}''  & 0    \\
``\texttt{and it's not ""putin's war"", it' russian war against Ukrainian HTTPURL}''  & 1      \\
``\texttt{@USER @USER Biden is the one funding fascists in Ukraine lol}''  & 0    \\
\bottomrule
\end{tabular}
\caption{Examples of false annotations (\binarylabelshort) for \dataset.}
\label{tab:exampleposts_false}
\end{table}

\subsection{Difficult Examples}

\Cref{tab:exampleposts_false} lists five example posts, where the content may be perceived as difficult to classify.

\begin{table}[h!]
\centering
\scriptsize
\renewcommand{\arraystretch}{1.3}
\begin{tabular}{p{6cm}P{0.5cm}}
\toprule
Posts      & \binarylabelshort                                                                                                                                                                                                                                                                                       \\
\midrule
``\texttt{@USER Oh, this may be a great read for you if your brain is so smooth that you truly believe everything wrong in the US political system is based solely off Russia. HTTPURL}''  & 1    \\
``\texttt{@USER when a nation is fascist they're fascist against the most vulnerable ones - women, minorities, disabled ppl so my point is that we should both acknowledge that most ppl here are fascists and support dissenters and not essentialism the fascism to the entire russian people}''  & 1    \\
``\texttt{@USER @USER @USER I do not support Russia. But the amount of nazis that are in Ukraine is staggering. If u paid attention to any news outlets the past ten years you would know the military and govt is infested with them. Don’t get me wrong, there are millions of normal good hearted…}''  & 1    \\
``\texttt{@USER Russian trolls can be trusted than your beloved media.}''  & 1      \\
``\texttt{Explain to me again how Russia is a threat to the United States. URL}''  & 0    \\
\bottomrule
\end{tabular}
\caption{Difficult examples and their annotation (\binarylabelshort) for \dataset.}
\label{tab:exampleposts_difficult}
\end{table}

\clearpage 

\section{Example Posts for \datasetplus}\label{sec:exampleposts}

\Cref{tab:exampleposts_i} and \Cref{tab:exampleposts_c} each list five example posts and their corresponding \binarylabel (\binarylabelshort) and \strategylabel (\strategylabelshort) for the two additional contexts in \datasetplus of \protect\circledorange{M} anti-Muslim propaganda in India and \protect\circledgreen{U} anti-Uyghur propaganda respectively.

\begin{table*}[h!]
\centering
\scriptsize
\renewcommand{\arraystretch}{1.3}
\begin{tabular}{p{12.9cm}P{0.5cm}p{1.42cm}}
\toprule
Posts                                                                                                                                                                                                                                                                                             & \binarylabelshort & \strategylabelshort                        \\
\midrule
``\texttt{Muslim couple killed in India over son 's relationship with Hindu girl HTTPURL}''                                                                                                                                  & False  & ---          \\
``\texttt{India faces a growing threat from Muslim children indoctrinated into the jihad ideology HTTPURL}''                                       & True  & Against main opposition            \\
``\texttt{BIG BREAKING :high\_voltage: :high\_voltage: \#IndianArmy has Conducted massive strike on \#Pakistan :Pakistan: based Terrorist trying to enter Kashmir at LOC . Army has downed several Jlhadi $\&$ recover big amount of Arms . It 's not Surgical Strike but yes a big Operation on Pakistan definitely :fire: IN :framed\_picture:  Representation HTTPURL}''                     & True  & Pro own stance     \\
``\texttt{Hindu India doing more to help \#Armenia AM and Greece GR then the so called Christian nations of the west The “ pro European ” EU is ofc silent as usual , too busy flooding \#Europe with Muslim migrants I guess HTTPURL}''                                                                                & True  & Against other oppositions  \\
``\texttt{Pakistanis are also ill minded people , they spill the salt on our wound and even justify the lynching , and make our fun by saying ' Thankyou Jinnah ' . HTTPURL}'' & True  & Other  \\
\bottomrule
\end{tabular}
\caption{\protect\circledorange{M} Example posts for the context of anti-Muslim propaganda in India. \binarylabelshort is a \binarylabel for whether a post is propaganda or not. \strategylabelshort is the \strategylabel.}
\label{tab:exampleposts_i}
\end{table*}

\begin{table*}[h!]
\centering
\scriptsize
\renewcommand{\arraystretch}{1.3}
\begin{tabular}{p{12.9cm}P{0.5cm}p{1.42cm}}
\toprule
Posts                                                                                                                                                                                                                                                                                             & \binarylabelshort & \strategylabelshort                        \\
\midrule
``\texttt{Stop Uyghur Genocide Stop xi jinping Stop fascist chinese government @USER \#UyghurGenocide HTTPURL}''                                                                                                                                  & False  & ---          \\
``\texttt{:fire: :fire: 1997 article : \#Uyghur Card to Break Up China :small\_blue\_diamond: West funded radical \#Uyghurs who wanted to blow up "" New Silk Road "" trains in \#Xinjiang :small\_blue\_diamond: "" Uyghur Liberation Front "" $= >$ offices in US / UK :small\_blue\_diamond: "" Turkestan "" $\&$ "" Uyghur genocide "" $= >$ talking points in 1994 :thread: 1 HTTPURL}''                                       & True  & Against main opposition            \\
``\texttt{:rainbow: :sun\_with\_face: Xinjiang people smile brightly , embracing life 's pleasures . :grinning\_squinting\_face: :sun: Contentment is the greatest wealth . \#Xinjiang \#happylife \#smiles HTTPURL}''                     & True  & Pro own stance     \\
``\texttt{\#UyghurGenocide itself is a part of Western propaganda , there is no concept of genocide among the Chinese people HTTPURL}''                                                                                & True  & Against other oppositions  \\
``\texttt{ON ENERGY ( TRANSITION ) Z A China : Do n't Let The New Deceive You . Save What You Already Have / Know Aggressive Europe : Do n't Use Coal , South Africa . It Is Bad For You . Send It All To Us Gwede Mantashe : The Chinese Said To Make Sure That You Strengthen To A Level Of Reliability , ... HTTPURL HTTPURL .}'' & True  & Other  \\
\bottomrule
\end{tabular}
\caption{\protect\circledgreen{U} Example posts for the context of anti-Uyghur propaganda. \binarylabelshort is a \binarylabel for whether a post is propaganda or not. \strategylabelshort is the \strategylabel.}
\label{tab:exampleposts_c}
\end{table*}

\section{Propaganda-Stance Label (\strategylabelshort)}\label{sec:psl}

Besides \binarylabelshort indicating wether a post is propagandistic or not, we also collect the propaganda-stance label (\strategylabelshort) for propagandistic posts, as detailed in \Cref{sec:data_annotation}. Workers assign one of the following four stances to propagandistic posts: (1)~against the main opposition (e.g., against Ukraine), (2)~pro own stance (e.g., pro Russian government), (3)~against other oppositions (e.g., against Western countries), or (4)~other. \Cref{tab:psldist_r} (\protect\circledblue{R}), \Cref{tab:psldist_i} (\protect\circledorange{M}), and \Cref{tab:psldist_c} (\protect\circledgreen{U})

\begin{table}[h!]
\centering
\scriptsize
\renewcommand{\arraystretch}{1.3}
\begin{tabular}{p{0.1cm}P{1.3cm}P{1.3cm}P{1.3cm}P{1.3cm}}
\toprule
 & against the main opposition & against other oppositions & pro own stance & other                      \\
\midrule
$N$ & 2266  & 1285 & 914 & 145          \\
\bottomrule
\end{tabular}
\caption{\protect\circledblue{R} Distribution of \strategylabelshort for the context of Russian propaganda.}
\label{tab:psldist_r}
\end{table}

\begin{table}[h!]
\centering
\scriptsize
\renewcommand{\arraystretch}{1.3}
\begin{tabular}{p{0.1cm}P{1.3cm}P{1.3cm}P{1.3cm}P{1.3cm}}
\toprule
 & against the main opposition & against other oppositions & pro own stance & other                      \\
\midrule
$N$ & 211  & 13 & 31 & 82          \\
\bottomrule
\end{tabular}
\caption{\protect\circledorange{M} Distribution of \strategylabelshort for the context of anti-Muslim propaganda in India.}
\label{tab:psldist_i}
\end{table}

\begin{table}[h!]
\centering
\scriptsize
\renewcommand{\arraystretch}{1.3}
\begin{tabular}{p{0.1cm}P{1.3cm}P{1.3cm}P{1.3cm}P{1.3cm}}
\toprule
 & against the main opposition & against other oppositions & pro own stance & other                      \\
\midrule
$N$ & 5  & 58 & 129 & 64          \\
\bottomrule
\end{tabular}
\caption{\protect\circledgreen{U} Distribution of \strategylabelshort for the context of anti-Uyghur propaganda.}
\label{tab:psldist_c}
\end{table}

\section{Propaganda Techniques}\label{sec:proptechniques}

\begin{figure*}[p]
    \captionsetup{position=top}
    \centering
    \includegraphics[width=0.95\linewidth]{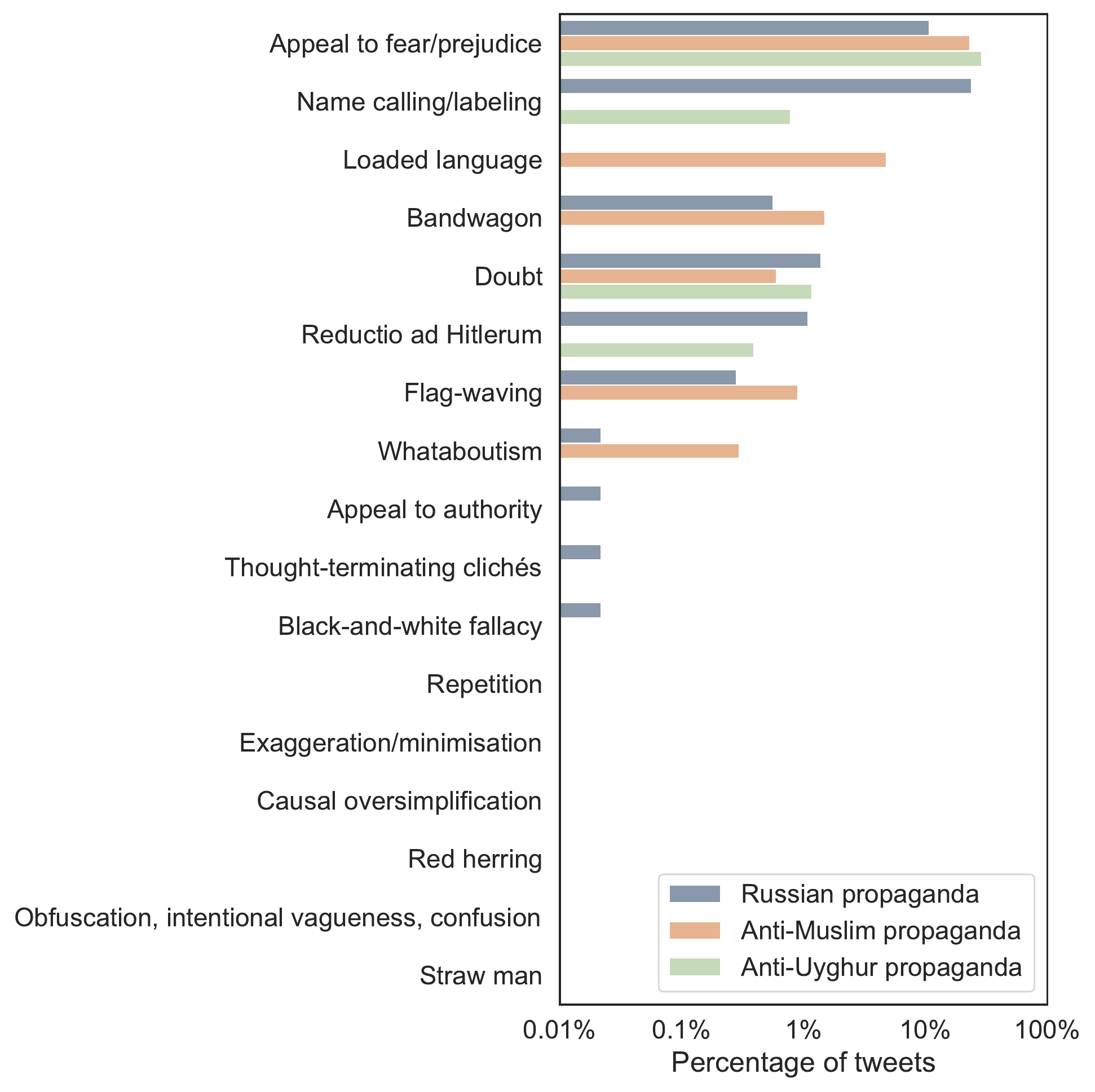}
    \caption{Percentage of the propaganda techniques of all propagandistic tweets for each context of the three contexts in \datasetplus: \protect\circledblue{R} Russian propaganda, \protect\circledorange{M} anti-Muslim propaganda in India, and \protect\circledgreen{U} anti-Uyghur propaganda.}
\label{fig:proptechniques}
\end{figure*}

We enrich both \dataset and \datasetplus with weak labels for the 18 propaganda techniques defined by \citet{DaSanMartino.2019}. We use a similar approach to \citet{Vijayaraghavan.2022} for weak labeling, i.e., we assign weak labels according to their mapping of propaganda phrases to propaganda techniques. In \Cref{fig:proptechniques}, we plot the distribution of the propaganda techniques for the three contexts of \protect\circledblue{R} Russian propaganda, \protect\circledorange{M} anti-Muslim propaganda in India, and \protect\circledgreen{U} anti-Uyghur propaganda.

\section{Linguistic Dimensions}\label{appendix:lingdim}

In \Cref{tab:lingpatterns_R} (\protect\circledblue{R}), \Cref{tab:lingpatterns_M} (\protect\circledorange{M}), and \Cref{tab:lingpatterns_U} (\protect\circledgreen{U}) we analyze linguistic dimensions to compare propagandistic vs. non-propagandistic posts. We use the LIWC2015 dictionary \cite{Pennebaker.2015} for the linguistic analysis. The measurements are based on the percentage of words corresponding to a specific linguistic dimension (e.g., anger) of the entire post. Sentiment is normalized to a range of $[-1, 1]$ for better comparability. We calculate the $p$-value based on Welch's $t$-test \cite{Welch.1947} for the means of the propaganda vs. no-propaganda class.

In the comparison between propagandistic and non-propagandistic posts in the context of Russian propaganda (\protect\circledblue{R}), some key differences emerge. Propagandistic posts tend to use more emotional language, including higher levels of anger ($p < 0.01$), anxiety ($p < 0.001$), and negative emotions ($p < 0.001$). They also focus more on future events ($p < 0.001$), whereas non-propagandistic posts put more emphasis on the past ($p < 0.001$). Propagandistic posts mention risk-related topics significantly more frequently ($p < 0.001$). Additionally, propagandistic posts include significantly more language that suggests insights ($p < 0.001$) and interrogation ($p < 0.01$). Interestingly, propagandistic posts have a tendency to use more negations (p < 0.001) but less certainty ($p < 0.05$) in their statements. Despite these differences, both propagandistic and non-propagandistic posts show similar levels of cognitive processing, social references, and swear words.

In the context of anti-Muslim propaganda in India (\protect\circledorange{M}), propagandistic posts barely differ from non-propagandistic posts in their linguistic patterns. Non-propagandistic posts exhibit more anger ($p < 0.05$), anxiety ($p < 0.001$), and negative emotions ($p < 0.001$) compared to propagandistic posts. In contrast, propagandistic posts feature higher levels of positive emotions ($p < 0.01$) and use more language indicating insights ($p < 0.05$). Both types of posts are similar in their overall emotional content, focus on time (past, present, future), and use of language related to certainty, negations, and social topics. Notably, the overall sentiment in non-propagandistic posts is more negative ($p < 0.01$).

The linguistic patterns in propagandistic vs. non-propagandistic posts in the context of anti-Uyghur propaganda show significant differences. On average, propagandistic posts have approximately three times more words linked to anxiety ($p < 0.001$) and 1.7 times more words linked to negative emotions ($p < 0.001$). Propagandistic posts exhibit a greater use of cognitive processes ($p < 0.001$), indicating more complex language use. They also tend to focus more on the future compared to non-propagandistic posts ($p < 0.01$). Additionally, these posts have a higher tendency to differentiate and use numbers ($p < 0.05$), pointing to a more detailed and specific style. In contrast, non-propagandistic posts exhibit more positive emotions ($p < 0.01$) and perception-related language ($p < 0.01$). The overall sentiment in propagandistic vs. non-propagandistic posts is significantly different ($p < 0.001$). On average, non-propagandistic posts are slightly positive while propagandistic posts are rather negative.

Overall, several consistent linguistic patterns emerge: Emotionally charged language, particularly negative emotions, is prominent in propagandistic posts. Propagandistic posts more often emphasize future events, suggesting a strategic intent to shape expectations or fears. Additionally, propagandistic posts often exhibit a more complex use of language, indicating a calculated approach to persuasion. These findings highlight the nuanced and multifaceted nature of propaganda across different scenarios.

\begin{table*}[h!]
\centering
\scriptsize
\begin{tabular}{lccccl}
\toprule
{} & \multicolumn{2}{c}{Propaganda} & \multicolumn{2}{c}{No Propaganda} &       \multirow{2}{*}{\emph{p}-Value} \\
\cmidrule(lr){2-3}\cmidrule(lr){4-5}
{} &       Mean &    SD &          Mean & SD & \\
\midrule
Affect              &      6.285 &  5.099 &         5.995 &  4.580 &   0.000 (***) \\
Anger               &      2.538 &  3.366 &         2.407 &  2.865 &    0.005 (**) \\
Anxiety             &      0.569 &  1.575 &         0.438 &  1.280 &   0.000 (***) \\
Causation           &      1.595 &  2.488 &         1.601 &  2.327 &  0.857 (n.s.) \\
Certainty           &      1.312 &  2.313 &         1.390 &  2.229 &     0.029 (*) \\
Cognitive processes &      8.554 &  6.508 &         8.463 &  5.768 &  0.334 (n.s.) \\
Comparison          &      1.772 &  2.823 &         1.495 &  2.416 &   0.000 (***) \\
Death               &      1.503 &  2.655 &         1.433 &  2.259 &  0.060 (n.s.) \\
Differentiation     &      2.431 &  3.227 &         2.600 &  3.027 &   0.001 (***) \\
Discrepancy         &      1.082 &  2.148 &         1.038 &  2.030 &  0.175 (n.s.) \\
Feeling             &      0.223 &  0.940 &         0.186 &  0.788 &    0.004 (**) \\
Future focus        &      0.864 &  1.997 &         0.737 &  1.685 &   0.000 (***) \\
Past focus          &      2.277 &  3.243 &         2.519 &  3.099 &   0.000 (***) \\
Present focus       &      8.191 &  5.681 &         8.297 &  5.203 &  0.212 (n.s.) \\
Health              &      0.358 &  1.284 &         0.275 &  1.067 &   0.000 (***) \\
Insight             &      1.571 &  2.538 &         1.365 &  2.133 &   0.000 (***) \\
Interrogation       &      1.227 &  2.198 &         1.322 &  2.091 &    0.005 (**) \\
Negation            &      1.474 &  2.442 &         1.653 &  2.340 &   0.000 (***) \\
Negative emotions   &      4.108 &  4.277 &         3.802 &  3.677 &   0.000 (***) \\
Numbers             &      0.517 &  1.504 &         0.452 &  1.299 &    0.002 (**) \\
Perception          &      1.598 &  2.658 &         1.437 &  2.286 &   0.000 (***) \\
Positive emotions   &      2.135 &  3.146 &         2.167 &  2.922 &  0.505 (n.s.) \\
Power               &      4.638 &  4.633 &         4.175 &  3.932 &   0.000 (***) \\
Quantifiers         &      1.489 &  2.397 &         1.484 &  2.306 &  0.881 (n.s.) \\
Risk                &      1.195 &  2.379 &         0.953 &  1.874 &   0.000 (***) \\
Sadness             &      0.322 &  1.210 &         0.249 &  0.894 &   0.000 (***) \\
Sentiment           &     -0.297 &  0.530 &        -0.315 &  0.529 &     0.033 (*) \\
Social              &      7.285 &  6.013 &         7.356 &  5.388 &  0.419 (n.s.) \\
Swear words         &      0.195 &  0.991 &         0.196 &  0.938 &  0.922 (n.s.) \\
Tentative           &      1.608 &  2.635 &         1.427 &  2.259 &   0.000 (***) \\
\bottomrule
\end{tabular}
\caption{\protect\circledblue{R} Linguistic dimension analysis for the context of Russian propaganda using the LIWC2015 dictionary. Measurements reflect the percentage of words in a post linked to the specific linguistic dimension. Sentiment is normalized to a range of $[-1, 1]$ for better comparability. The $p$-values are based on the Welch's $t$-test. Significance levels: ***: $p<0.001$, **: $p<0.01$, *: $p<0.05$, and n.s.: not significant.}
\label{tab:lingpatterns_R}
\end{table*}

\begin{table*}[h!]
\centering
\scriptsize
\begin{tabular}{lccccl}
\toprule
{} & \multicolumn{2}{c}{Propaganda} & \multicolumn{2}{c}{No Propaganda} &       \multirow{2}{*}{\emph{p}-Value} \\
\cmidrule(lr){2-3}\cmidrule(lr){4-5}
{} &       Mean &    SD &          Mean & SD & \\
\midrule
Affect              &      5.133 &  4.460 &         5.147 &  4.141 &  0.960 (n.s.) \\
Anger               &      1.520 &  2.395 &         2.121 &  2.768 &   0.001 (***) \\
Anxiety             &      0.342 &  1.215 &         0.726 &  2.295 &    0.004 (**) \\
Causation           &      1.154 &  2.276 &         1.264 &  2.028 &  0.435 (n.s.) \\
Certainty           &      0.907 &  1.947 &         0.801 &  1.777 &  0.390 (n.s.) \\
Cognitive processes &      6.193 &  6.204 &         5.762 &  5.714 &  0.275 (n.s.) \\
Comparison          &      1.418 &  2.446 &         1.295 &  2.080 &  0.408 (n.s.) \\
Death               &      0.304 &  1.288 &         0.556 &  1.489 &    0.009 (**) \\
Differentiation     &      1.822 &  2.810 &         1.889 &  3.434 &  0.759 (n.s.) \\
Discrepancy         &      0.703 &  1.782 &         0.599 &  1.516 &  0.335 (n.s.) \\
Feeling             &      0.188 &  0.996 &         0.154 &  0.741 &  0.549 (n.s.) \\
Future focus        &      0.897 &  1.709 &         0.871 &  1.548 &  0.812 (n.s.) \\
Past focus          &      2.316 &  2.787 &         2.346 &  2.740 &  0.871 (n.s.) \\
Present focus       &      6.670 &  5.193 &         6.443 &  5.288 &  0.520 (n.s.) \\
Health              &      0.256 &  1.066 &         0.226 &  0.873 &  0.628 (n.s.) \\
Insight             &      1.295 &  2.399 &         0.948 &  1.931 &     0.014 (*) \\
Interrogation       &      1.457 &  2.083 &         1.433 &  1.868 &  0.852 (n.s.) \\
Negation            &      1.213 &  2.483 &         0.965 &  1.917 &  0.081 (n.s.) \\
Negative emotions   &      2.881 &  3.328 &         3.402 &  3.359 &     0.021 (*) \\
Numbers             &      0.727 &  1.529 &         0.886 &  1.538 &  0.123 (n.s.) \\
Perception          &      1.666 &  2.641 &         1.424 &  2.387 &  0.144 (n.s.) \\
Positive emotions   &      2.231 &  3.040 &         1.732 &  2.648 &    0.008 (**) \\
Power               &      3.372 &  3.657 &         3.491 &  4.094 &  0.656 (n.s.) \\
Quantifiers         &      1.365 &  2.214 &         1.347 &  2.649 &  0.914 (n.s.) \\
Risk                &      0.507 &  1.396 &         0.898 &  2.069 &    0.002 (**) \\
Sadness             &      0.338 &  1.086 &         0.241 &  0.832 &  0.116 (n.s.) \\
Sentiment           &     -0.137 &  0.520 &        -0.248 &  0.511 &    0.001 (**) \\
Social              &      7.097 &  5.797 &         7.913 &  6.420 &  0.051 (n.s.) \\
Swear words         &      0.049 &  0.492 &         0.043 &  0.482 &  0.835 (n.s.) \\
Tentative           &      0.959 &  1.981 &         1.018 &  2.062 &  0.664 (n.s.) \\
\bottomrule
\end{tabular}
\caption{\protect\circledorange{M} Linguistic dimension analysis for the context of anti-Muslim propaganda in India using the LIWC2015 dictionary. Measurements reflect the percentage of words in a post linked to the specific linguistic dimension. Sentiment is normalized to a range of $[-1, 1]$ for better comparability. The $p$-values are based on the Welch's $t$-test. Significance levels: ***: $p<0.001$, **: $p<0.01$, *: $p<0.05$, and n.s.: not significant.}
\label{tab:lingpatterns_M}
\end{table*}

\begin{table*}[h!]
\centering
\scriptsize
\begin{tabular}{lccccl}
\toprule
{} & \multicolumn{2}{c}{Propaganda} & \multicolumn{2}{c}{No Propaganda} &       \multirow{2}{*}{\emph{p}-Value} \\
\cmidrule(lr){2-3}\cmidrule(lr){4-5}
{} &       Mean &    SD &          Mean & SD & \\
\midrule
Affect              &      4.760 &  4.636 &         4.820 &  4.875 &  0.864 (n.s.) \\
Anger               &      1.117 &  2.246 &         0.762 &  1.794 &     0.011 (*) \\
Anxiety             &      0.449 &  1.590 &         0.159 &  0.725 &   0.000 (***) \\
Causation           &      1.451 &  2.594 &         1.302 &  2.225 &  0.375 (n.s.) \\
Certainty           &      1.363 &  2.507 &         1.156 &  2.039 &  0.187 (n.s.) \\
Cognitive processes &      7.787 &  6.409 &         6.320 &  5.802 &   0.001 (***) \\
Comparison          &      1.509 &  2.594 &         1.336 &  2.594 &  0.360 (n.s.) \\
Death               &      0.606 &  1.644 &         0.447 &  1.275 &  0.113 (n.s.) \\
Differentiation     &      2.249 &  3.110 &         1.652 &  2.709 &    0.004 (**) \\
Discrepancy         &      0.959 &  2.329 &         0.753 &  1.795 &  0.144 (n.s.) \\
Feeling             &      0.203 &  0.909 &         0.321 &  1.068 &  0.113 (n.s.) \\
Future focus        &      0.867 &  1.970 &         0.531 &  1.425 &    0.004 (**) \\
Past focus          &      2.140 &  3.631 &         1.643 &  2.499 &     0.016 (*) \\
Present focus       &      7.441 &  5.772 &         6.986 &  5.468 &  0.258 (n.s.) \\
Health              &      0.437 &  1.459 &         0.592 &  1.733 &  0.200 (n.s.) \\
Insight             &      1.375 &  2.607 &         1.179 &  2.356 &  0.265 (n.s.) \\
Interrogation       &      1.118 &  2.298 &         0.892 &  1.712 &  0.097 (n.s.) \\
Negation            &      1.184 &  2.259 &         1.028 &  1.983 &  0.295 (n.s.) \\
Negative emotions   &      2.568 &  3.611 &         1.562 &  2.670 &   0.000 (***) \\
Numbers             &      0.562 &  1.541 &         0.375 &  1.048 &     0.031 (*) \\
Perception          &      1.516 &  2.603 &         2.220 &  3.421 &    0.003 (**) \\
Positive emotions   &      2.180 &  3.131 &         3.207 &  4.649 &    0.001 (**) \\
Power               &      3.569 &  4.077 &         3.118 &  3.765 &  0.107 (n.s.) \\
Quantifiers         &      1.302 &  2.341 &         1.392 &  2.304 &  0.594 (n.s.) \\
Risk                &      0.922 &  2.269 &         0.548 &  1.346 &    0.002 (**) \\
Sadness             &      0.302 &  1.132 &         0.234 &  0.944 &  0.344 (n.s.) \\
Sentiment           &     -0.078 &  0.516 &         0.077 &  0.573 &   0.000 (***) \\
Social              &      7.226 &  6.197 &         6.644 &  6.000 &  0.185 (n.s.) \\
Swear words         &      0.113 &  0.879 &         0.087 &  0.533 &  0.575 (n.s.) \\
Tentative           &      1.370 &  2.660 &         0.987 &  1.837 &     0.011 (*) \\
\bottomrule
\end{tabular}
\caption{\protect\circledgreen{U} Linguistic dimension analysis for the context of anti-Uyghur propaganda using the LIWC2015 dictionary. Measurements reflect the percentage of words in a post linked to the specific linguistic dimension. Sentiment is normalized to a range of $[-1, 1]$ for better comparability. The $p$-values are based on the Welch's $t$-test. Significance levels: ***: $p<0.001$, **: $p<0.01$, *: $p<0.05$, and n.s.: not significant.}
\label{tab:lingpatterns_U}
\end{table*}

\section{Implemenation Details for Full Fine-tuning} \label{sec:details-finetuning}

For fine-tuning, we add a linear layer to the hidden representation of the $\mathrm{[CLS]}$ token. The PLMs are then fine-tuned using the transformer framework from Huggingface \cite{Wolf.2020}. We set the maximum sequence length to 128. We use a training batch size of 32 and a learning rate of \mbox{4e-5}. We freeze the first 16 layers of the PLMs. For BERT-large and RoBERTa-large, we add emoji-tokens to the vocabulary due to their frequent and meaningful use in social media. For BERTweet-large, emoji-tokens were already incorporated in the vocabulary during training. The number of parameters is 340 M, 355 M, and 355 M for the PLMs BERT-large, RoBERTa-large, and BERTweet-large, respectively. Weight updates are performed using the AdamW-optimizer \cite{Loshchilov.2019}. We fine-tune for a maximum number of 5 epochs. We validate the performance every 500 steps for fine-tuning with TWEETSPIN and every 50 steps otherwise. Early stopping is used when the loss on the validation set does not decrease for more than 5 validation steps.

All experiments are conducted on a Ubuntu 20.04 system, with 2.30 GHz Intel Xeon Silver 4316 CPU and two NVIDIA A100-PCIE-40GB GPUs.

\section{Prompt-Based Learning}\label{sec:pbl_appendix}

In the following, we report implementation details and further experimental results of prompt-based learning on our datasets \dataset and \datasetplus.

\subsection{Implementation Details for Prompt-Based Learning}

In our implementation of LM-BFF and LM-BFF-AT, we use the OpenPromt framework \cite{Ding.2022}. For template generation, we choose an initial verbalizer with label words ``\texttt{propaganda}'' (propaganda) and ``\texttt{truth}'' (no propaganda) and a cloze prompt format \cite{Liu.2023}. We choose T5-large ($\sim$770 M parameters) for generating candidates for (i) automatic template generation. We choose RoBERTa-large \cite{Liu.2019} as the underlying PLM for (ii) automatic verbalizer generation and (iii) prompt-based fine-tuning due to its overall superior performance. We freeze the first 16 layers to control for overfitting and choose a learning rate of 4e$-$5. We train for 50 epochs and choose the best checkpoint. We set the batch size depending on $k'$; see \Cref{tab:batch-sizes-pl}. For all other hyper-parameters, we choose the same as those presented in \citet{Gao.2021}.

Again, all experiments are conducted on a Ubuntu 20.04 system, with 2.30 GHz Intel Xeon Silver 4316 CPU and two NVIDIA A100-PCIE-40GB GPUs.

\begin{table}[h!]
\scriptsize
\centering
\begin{tabular}{lcccc}
\toprule
           & $k' = 16$ & $k' = 32$ & $k' = 64$ & $k' = 128$  \\
\midrule
Batch size & 4  & 8  & 16 & 32  \\
\bottomrule
\end{tabular}
\caption{Batch size of prompt-based learning for different numbers of overall samples ($k = 4 \times k'$).}
    \label{tab:batch-sizes-pl}
\vspace{-.5cm}
\end{table}

\subsection{Hyper-Parameters for Classification Heads of LM-BFF-AT}\label{sec:hplmbffat}

For our extension of the LM-BFF method, namely LM-BFF-AT, we perform hyper-parameter tuning using grid search for the two classification heads, i.e., the elastic net and the neural net. The tuning grids are reported in \Cref{tab:grids-plclassification}. We implement the elastic net using Python's scikit-learn module. The neural net is implemented using PyTorch.

\begin{table}[h!]
\centering
\scriptsize
\begin{tabular}{p{1.5cm}p{2.2cm}p{2.9cm}}
\toprule
Classification head          & Hyper-parameter         & Grid                                 \\
\midrule
\multirow{2}{*}{Elastic net} & Cost                       & \{0.1, 0.25, 0.5, 1, 2, 4, 8\}             \\
                             & L1-ratio                & \{0.1, 0.15, 0.2, 0.25, 0.3, 0.35\}  \\
                             &                         &                                      \\
                             \midrule
\multirow{4}{*}{Neural net}  & Dropout                 & \{0.2, 0.4, 0.6\}                    \\
                             & Learning rate           & \{0.001, 0.01, 0.02\}                \\
                             & Batch size              & \{2, 4, 8, 16\}                             \\
                             & Neurons in hidden layer & $\mathit{input\_dim}*$\{0.5, 1, 1.5, 2\}                 \\
\bottomrule
\end{tabular}
\caption{Grids for hyper-parameter tuning of the two classification heads for prompt-based learning with LM-BFF-AT. The cost $C = \frac{1}{\lambda}$ is the inverse of the regularization strength. L1-ratio is the elastic net mixing parameter (i.e., is equivalent to only using an L2-penalty (or L1-penalty) for L1-ratio$=0$ (or L1-ratio$=1$) and mixes both for $0 < $ L1-ratio $< 1$). Here, $\mathit{input\_dim}$ is the number of input features and therefore depends on whether only probabilities from the verbalizers or also author representations are used.}
\label{tab:grids-plclassification}
\end{table}

\subsection{Evaluation of the Auxiliary Task in LM-BFF-AT}\label{sec:at-performance}

In \Cref{tab:performance-at}, we report the performance of the auxiliary task in LM-BFF-AT, i.e., the performance of prompt-based learning using \strategylabelshort on \dataset. 

\begin{table}[h!]
\centering
\scriptsize
\renewcommand{\arraystretch}{1.3}
\begin{tabular}{p{0.2cm}P{1cm}P{1cm}P{1cm}}
\toprule
 $k'$ & Weighted Precision & Weighted Recall &  Weighted F1 \\
\midrule
 16 &       75.06 (2.54) &    66.17 (6.81) & 69.84 (4.02) \\
 32 &        75.6 (1.21) &   64.93 (10.73) & 68.75 (7.48) \\
 64 &       77.46 (2.19) &    75.31 (4.45) & 76.13 (2.15) \\
128 &       77.18 (3.45) &    74.39 (3.87) & 75.41 (1.86) \\
\bottomrule
\multicolumn{4}{l}{Stated: mean (SD)}
\end{tabular}
\caption{Evaluation results for the auxiliary task of prompt-based learning using \strategylabelshort on \dataset.}
\label{tab:performance-at}
\end{table}

\subsection{Baselines for Prompt-Based Learning} \label{sec:exp-fs-baselines}

In \Cref{tab:performance-fs-baselines} we evaluate the following three baselines for prompt-based learning: (i)~A manual template with a manual verbalizer (FixedT + FixedV), (ii)~a manual template with automatic verbalizer generation (AutoV), and (iii)~automatic template generation and a manual verbalizer (AutoT). For the manual template we select ``\texttt{This is the} $\mathrm{[MASK]}$''. The manual verbalizer maps ``\texttt{propaganda}'' to the class of propaganda and ``\texttt{truth}'' to the class of no propaganda. We show the results for $k'=16$ and report the mean (and standard deviation) over five runs for each baseline. Overall, the performance of the baselines is inferior to the methods LMBFF and LMBFF-AT in \Cref{sec:exp-fs}.

\begin{table}[h!]
\centering
\scriptsize
\renewcommand{\arraystretch}{1.3}
\begin{tabular}{p{1.7cm}P{0.9cm}P{0.9cm}P{0.9cm}P{0.9cm}}
\toprule
         Method &    Precision &        Recall &           F1 &          AUC \\
\midrule
FixedT + FixedV & 21.98 (4.36) & 56.71 (22.26) & 30.06 (2.61) & 64.34 (4.83) \\
    LMBFF-autoV & 20.51 (1.24) &  69.06 (4.77) & 31.59 (1.54) &  64.18 (2.60) \\
    LMBFF-autoT &  19.45 (1.70) & 59.96 (19.37) & 28.77 (3.65) & 59.85 (4.64) \\
\bottomrule
\multicolumn{4}{l}{Stated: mean (SD).}
\end{tabular}
\caption{Evaluation results for the baselines of prompt-based learning on \dataset for $k'=16$. We evaluate (i)~a manual template with a manual verbalizer (FixedT + FixedV), (ii)~a manual template with automatic verbalizer generation (AutoV), and (iii)~automatic template generation and a manual verbalizer (AutoT).}
\label{tab:performance-fs-baselines}
\end{table}

\subsection{Prompt-Based Learning with Increasing Number of Few-Shot Samples} \label{sec:exp-fs-higherk}

\begin{table}[h!]
\centering
\scriptsize
\renewcommand{\arraystretch}{1.3}
\begin{tabular}{p{1.7cm}P{0.9cm}P{0.9cm}P{0.9cm}P{0.9cm}}
\toprule
         $k'$ &    Precision &        Recall &           F1 &          AUC \\
\midrule
256 & 35.42 (1.89) & 82.14 (2.99) & 49.43 (1.36) & 85.47 (0.76) \\
    512 & 38.49 (0.58) &  80.99 (0.76) & 52.18 (0.57) &  86.51 (0.22) \\
\bottomrule
\multicolumn{4}{l}{Stated: mean (SD).}
\end{tabular}
\caption{Evaluation results for prompt-based learning on \dataset with increasing number of few-shot samples $k'$.}
\label{tab:performance-fs-higherk}
\end{table}

We also conduct experiments with larger numbers of few-shot samples, namely, for $k'=256$ ($k=1024$) and $k'=512$ ($k=2048$). We report the results in \Cref{tab:performance-fs-higherk}. Overall, we observe that the performance still increases with $k'$; however, the increase is no longer linear as opposed to our findings in \Cref{sec:exp-fs}. Specifically, we observe a diminishing gain in performance for $k'=512$. Our explanation for these findings is the following: prompt-based learning makes use of the masked language modeling (MLM) task of the underlying PLM for classification. Therefore, no new parameters need to be introduced for classification, which makes it beneficial to learn for low-resource settings. However, the benefits of lower parametrization diminish with higher $k'$ as more data is available for training potential new parameters. With larger amounts of annotated data, the introduction of new parameters specifically for the classification task becomes beneficial.

\section{Propaganda Detection with Additional Meta Information} \label{sec:exp-mf}

We extend our propaganda detection so that we not only use the content but the additional meta information (i.e., author features and pinned-post features) for propaganda detection.

\subsection{Meta Information} 


We enrich \dataset with additional meta information from the social network. Here, we use a comprehensive set of author features (e.g., number of followers, account age, verified status) and pinned-post\footnote{Every Twitter/X user can choose to pin one (self-written) post to her/his account, which is then always displayed at the top of the profile.} features (e.g., post age, number of likes, number of reposts). Our data further includes the profile description of authors using embeddings from SBERT \cite{Reimers.2019}. We choose SBERT due to its strength in capturing meaningful representations from short text \cite{Reimers.2019}. A few posts corresponded to authors whose accounts were already deleted, which reduces our dataset to the final size of $N=\num{29596}$. We note that for fair benchmarking we used the dataset of size $N=\num{29596}$ for all experiments. The full list of features is in \Cref{appendix:summary_stats}. 

\subsection{Summary Statistics}\label{appendix:summary_stats}

\begin{table*}[t!]
\centering
\scriptsize
\begin{tabular}{p{5cm}P{2cm}P{2cm}P{2cm}P{2cm}} 
\toprule
                                 & \multicolumn{2}{c}{Propaganda = true}                     & \multicolumn{2}{c}{Propaganda = false}         \\
                                 \cmidrule(lr){2-3}\cmidrule(lr){4-5}
                                 & \multicolumn{1}{c}{Mean} & \multicolumn{1}{c}{Median} & \multicolumn{1}{c}{Mean} & \multicolumn{1}{c}{Median} \\ 
\toprule
Verified (=1; not=0)                         & 0.09                 & 0                & 0.19                 & 0                \\
\#Followers                        & 102784.72            & 513      & 595308.01            & 13405      \\
\#Following                        & 1927.58              & 530.5            & 3237.82              & 786            \\
\#Posts                           & 37780.17             & 11123.5            & 64060.48             & 16811          \\
\#Listed$^\ast$                          & 633.36               & 5             & 2449.10              & 11            \\
Account age (in days)            & 2272.17              & 1938             & 2763.28              & 2895             \\
\#Followers divided by Account age        & 28.05                & 0.37              & 161.74               & 0.71            \\
\#Following divided by Account age        & 1.19                 & 0.34                & 1.48                 & 0.34                \\
\#Posts divided by Account age           & 19.34                & 7.915               & 25.76                & 8.58               \\

\bottomrule
\multicolumn{4}{l}{$^\ast$Number of Twitter/X lists (i.e., a curated group of accounts) comprising the account of the author.}
\end{tabular}
\caption{Summary statistics of author features in \dataset.}
\label{tab:sum-stats-af}
\end{table*}

\begin{table*}[t!]
\centering
\scriptsize
\begin{tabular}{p{5cm}P{2cm}P{2cm}P{2cm}P{2cm}} 
\toprule
                                 & \multicolumn{2}{c}{Propaganda = true}                     & \multicolumn{2}{c}{Propaganda = false}         \\
                                 \cmidrule(lr){2-3}\cmidrule(lr){4-5}
                                 & \multicolumn{1}{c}{Mean} & \multicolumn{1}{c}{Median} & \multicolumn{1}{c}{Mean} & \multicolumn{1}{c}{Median} \\ 
\toprule
Account age (in days)                    & 365.30               & 198              & 366.82               & 170              \\
\#Reposts                         & 573.12               & 13             & 1051.95              & 17             \\
\#Replies                          & 251.85               & 6             & 416.71               & 7             \\
\#Likes                            & 2232.50              & 39            & 4336.85              & 62             \\
\#Quotes                           & 70.19                & 2              & 131.85               & 2            \\
\#Reposts divided by account age                   & 16.70                & 0.08              & 30.49                & 0.14              \\
\#Replies

divided by account age                  & 5.09                 & 0.03               & 11.95                & 0.05             \\
\#Likes divided by account age                      & 59.17                & 0.25              & 111.47               & 0.5              \\
\#Quotes divided by account age                     & 1.14                 & 0.01                & 3.18                 & 0.01               \\
\bottomrule
\end{tabular}
\caption{Summary statistics for pinned-post features of authors in \dataset.}
\label{tab:sum-stats-pt}
\end{table*}

\Cref{tab:sum-stats-af} reports summary statistics of the author features for \dataset. We compare mean and standard deviation separately for both propagandistic and non-propagandistic content. In line with previous findings \cite{Geissler.2022}, we find that authors of propagandistic content are, on average, less often verified, have fewer followers and posts, and are characterized by a younger account age. \Cref{tab:sum-stats-pt} reports summary statistics for the pinned-post features of the authors in \dataset.

\subsection{Adaptation of Methods} 

\textbf{(i)~Full fine-tuning:} We modify the classification head so that we perform full fine-tuning with author and pinned-post features. The self-description of post authors is encoded into a 768-dimensional vector using SBERT. We normalize the numerical features from \Cref{tab:sum-stats-af} and \Cref{tab:sum-stats-pt} and append them to the self-description vector to get an author representation. We concatenate the hidden representation of the $\mathrm{[CLS]}$ token generated by the PLM and the author representation and again feed them to a linear layer. The fine-tuning procedure and the hyper-parameters are identical to those described in \Cref{sec:details-finetuning}.

\textbf{(ii)~Prompt-based learning:} We create the same author representation as in (i). Here, we concatenate the author representation with the verbalizer probabilities for both \binarylabelshort and \strategylabelshort and again feed them to a classification head as in LM-BFF-AT. The prompt-based learning procedure is identical to that in \Cref{sec:promptlearning}, i.e., we select the optimal template and verbalizer from each run in \Cref{sec:exp-fs}. The tuning grids for hyper-parameter tuning for the two classification heads are identical to those in \Cref{tab:grids-plclassification}.

\subsection{Results} 

\Cref{tab:performance-MI} reports the prediction performance when additionally using the author representations. Values are in bold if the model that uses author representations outperforms the counterpart without them. Overall, we observe a tendency that the results improve when additionally using author representations. This tendency is seen for both full fine-tuning and prompt-based learning. However, the performance gain from  using the content is larger than the performance gain from using author representations. This can be expected as propaganda spreaders typically do not explicitly disclose their manipulative intention but instead aim to deceive users.

\begin{table}[t]
\centering
\scriptsize
\renewcommand{\arraystretch}{1.3}
\begin{tabular}{p{0.2cm}p{0.8cm}p{0.5cm}p{0.5cm}p{0.5cm}p{0.5cm}}
\toprule
          & PLM &            P &            R &           F1 &          AUC \\
\midrule
\multirow{5}{*}{\rotatebox[origin=c]{90}{(i) Full fine-tuning}}        & BERT & \textbf{64.58} (3.19) &  62.40 (1.94) & \textbf{63.42} (1.64) & \textbf{88.54} (0.81) \\
     & RoBERTa & \textbf{67.61} (3.05) & 69.59 (3.02) & 68.52 (2.04) & 91.37 (1.02) \\
    & BERTweet & 67.06 (3.39) & \textbf{72.68} (1.54) & \textbf{69.71} (1.79) & 92.06 (0.86) \\
    \midrule
    & $k'$ & P & R & F1 & AUC \\
    \midrule
\multirow{7}{*}{\rotatebox[origin=c]{90}{(ii) Prompt-based learning}}  & 16 & 22.38 (3.37) &  \textbf{64.94} (9.79) & \textbf{32.98} (3.49) & 64.99 (4.16) \\
    & 32 & \textbf{24.75} (6.24) &  65.16 (3.6) & \textbf{35.70} (4.16) & 69.29 (4.48) \\
    & 64 & \textbf{27.73} (2.80) &  70.72 (4.07) & \textbf{39.74} (2.88) & \textbf{74.88} (2.21) \\
    & 128 & 30.81 (2.86) &  76.38 (1.52) & 43.86 (3.02) & 79.46 (2.89) \\
\bottomrule
\multicolumn{5}{l}{Stated: mean (SD). P: precision. R: recall.}
\end{tabular}
\caption{Evaluation results of (i) full fine-tuning and (ii) prompt-based learning on \dataset while incorporating author representations. Results are shown in \textbf{bold} if the performance with author representations is better than the performance without them.}
\label{tab:performance-MI}
\end{table}

\clearpage

\end{document}